%% file: neurips_main.tex
\documentclass{article}

\usepackage[final,main]{neurips_2026}

\usepackage[utf8]{inputenc}
\usepackage[T1]{fontenc}
\usepackage{hyperref}
\usepackage{url}
\usepackage{booktabs}
\usepackage{amsfonts}
\usepackage{nicefrac}
\usepackage{microtype}
\usepackage{xcolor}
\usepackage{algorithm}
\usepackage{algpseudocode}
\usepackage{amsmath}
\usepackage{amssymb}
\usepackage{amsthm}
\usepackage{tikz}
\usetikzlibrary{arrows.meta, positioning, calc}
\usepackage{float}
\usepackage{tabularray}
\usepackage{enumitem}
\usepackage{longtable}
\usepackage{sidecap}
\usepackage{caption}

\hypersetup{hidelinks}

\newtheorem{theorem}{Theorem}
\newtheorem{lemma}[theorem]{Lemma}
\newtheorem{corollary}[theorem]{Corollary}

\title{Perturb and Correct: Post-Hoc Ensembles using Affine Redundancy}

\author{
  Eleanor Quint\\
  Department of Computing\\
  University of Nebraska--Lincoln\\
  Lincoln, NE 68508, USA\\
  \texttt{equint@cse.unl.edu}
}

\begin{document}

\maketitle

\begin{abstract}
Models that are nearly indistinguishable on in-distribution data can behave very differently under distribution shift. We introduce Perturb-and-Correct (P\&C), a post-hoc method for constructing epistemically diverse predictors from a single pretrained network. P\&C applies random hidden-layer perturbations with a least-squares correction in the following affine layer, producing predictors that agree on calibration data while remaining free to disagree away from it. We analyze this mechanism through the post-correction residual that remains in the activation of the layer corrected using calibration data, yielding a leverage-based preservation bound and a covariance interpretation of ensemble disagreement. Empirically, P\&C achieves a strong ID/OOD tradeoff across MuJoCo dynamics prediction and CIFAR-10 OOD detection, and scales to a pretrained ViT-B/16 on ImageNet-1K. In matched comparisons with Deep Ensembles, P\&C substantially reduces construction and storage cost. Our findings highlight the potential in further exploiting overparameterization as a strength of deep learning models.
\end{abstract}

\input{introduction}
\input{related_work}
\input{algorithm}
\input{theory}
\input{experiments}

\section{Conclusion}

We introduced Perturb-and-Correct (P\&C), a post-hoc method for constructing epistemically diverse ensembles from a single pretrained network. By pairing random hidden-layer perturbations with local affine correction, P\&C preserves in-distribution behavior while allowing disagreement to emerge where the same behavior fails to transfer. Our analysis makes this mechanism explicit through the exact post-correction residual: calibration changes control for part of the perturbation effect, while the remaining transfer defect controls the output-visible change and the resulting ensemble covariance. Across MuJoCo dynamics prediction, CIFAR-10 OOD detection, and ImageNet-1K with a pretrained ViT-B/16, P\&C preserves ID behavior while exposing useful variation under shift; an additional DistilBERT experiment extends the same construction to text. Its matched advantage over Deep Ensembles lies primarily in inexpensive post-hoc construction and compact storage, while member-dependent inference remains $O(M)$ and arbitrary distribution shift need not produce useful disagreement. Within these limits, P\&C provides a simple way to expose predictive degrees of freedom left unresolved after preserving a completed model on calibration data.

\begin{ack}
I would like to acknowledge the support of my partner. This work was not funded.

\textbf{Competing interests:} Within the relevant 36-month period, I was employed by Phaidra.
\end{ack}

\clearpage
\newpage

\bibliographystyle{plainnat}
\bibliography{main}

\clearpage
\appendix

\input{appendix_theory}
\input{appendix_conv}

\input{appendix_mujoco}
\input{appendix_cifar}
\input{appendix_imagenet}
\input{appendix_distilbert}

\end{document}

%% file: introduction.tex
\section{Introduction}

Modern neural networks are massively overparameterized. Many distinct parameter settings can realize models that are nearly indistinguishable on the data on which they are trained~\citep{d2022underspecification}. This flexibility forms a foundation of the modern deep learning regime, supporting effective first-order optimization~\citep{dauphin2014identifying,choromanska2015loss} and generalization phenomena distinct from less parameterized models, such as double descent~\citep{belkin2019reconciling}. At the same time, it contributes to a broader problem of \emph{underspecification}, where a training pipeline can return predictors with essentially identical in-distribution (ID) performance even though those predictors may behave very differently under distribution shift~\citep{d2022underspecification}. In such settings, the observed data do not fully determine model behavior away from their support, making epistemic uncertainty and out-of-distribution detection central to reliable deployment.

\begin{figure}[!t]
    \centering
    \includegraphics[width=\linewidth]{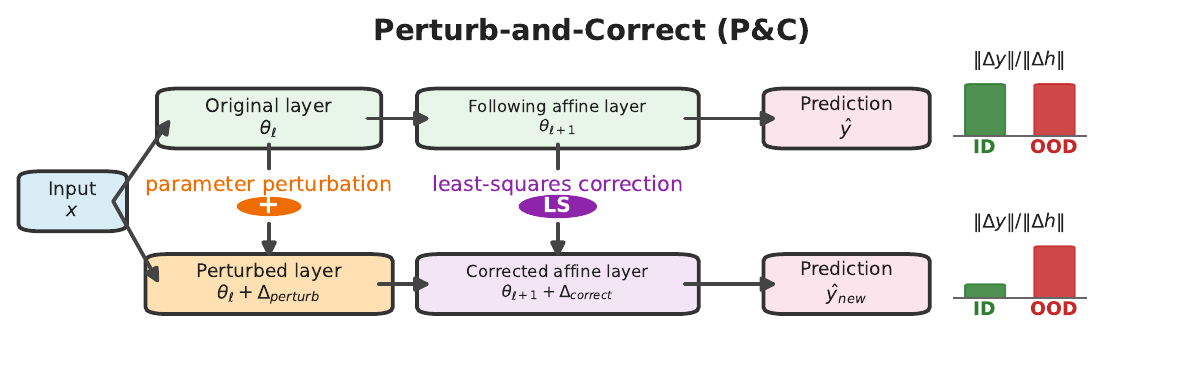}
    \caption{\textbf{Perturb-and-Correct.} A hidden-layer perturbation changes the representation $h$ to $h_v$. The following affine layer is refitted on ID calibration examples so that the corrected predictor remains close to the base model there. At a shifted input, the representation change may not be reconstructed from the changes observed on calibration data, leaving a post-correction residual that can produce ensemble disagreement.}
    \label{fig:PnC_schematic}
\end{figure}

In this work, we introduce \emph{Perturb-and-Correct} (P\&C), a post-hoc ensemble construction for exposing this unresolved model variation (Figure~\ref{fig:PnC_schematic}). P\&C samples random hidden-layer parameter perturbations and then fits a compensating correction in the following affine layer to reproduce the base model's next-layer pre-activations on ID calibration data. The correction suppresses perturbation effects that can be absorbed by the calibration geometry while leaving members free to differ where that repair no longer transfers. In this way, P\&C turns a single completed model into a family of nonlinear predictors that agree where calibration data constrain behavior and can disagree where they do not.

Our theory formalizes this intuition by analyzing what remains after a hidden-layer perturbation is corrected. We derive an exact post-correction residual in terms of the nonlinear representation changes induced by the applied perturbation. The residual separates the test-point change from the component reconstructed from calibration changes, yielding a leverage-based preservation bound and a \emph{transfer defect}: the portion of the perturbation effect that the calibration geometry fails to reconstruct. Near well-covered calibration representations, the repair can continue to suppress the perturbation; away from that geometry, the transfer defect can remain and produce disagreement. Random P\&C therefore defines a calibration-dependent random function whose covariance records shared, unresolved correction failures. The corrected-sensitivity expression is recovered by differentiating this exact object at the unperturbed model, while the exact residual describes the perturbation scales used by the ensemble.

\begin{figure}[t]
    \centering
    \includegraphics[width=\linewidth]{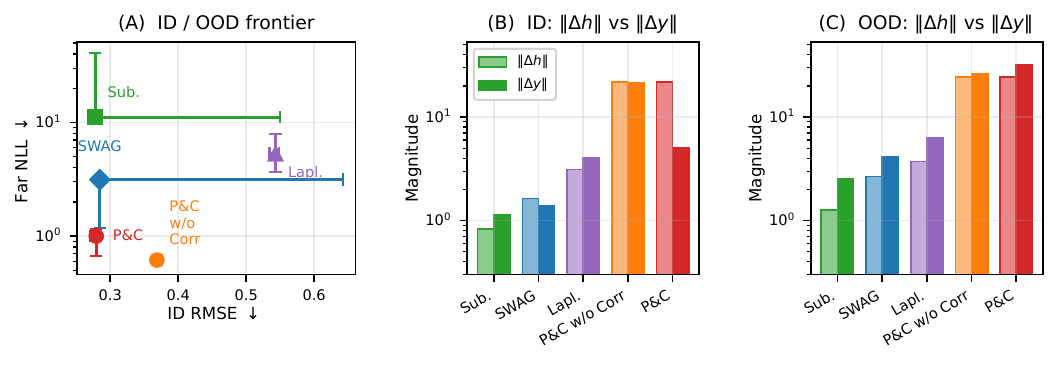}
    \caption{\textbf{P\&C improves the ID/OOD tradeoff on Ant-v5 by preserving large hidden perturbations while reducing their ID-visible effect.} \textbf{(A)} ID RMSE versus Far-OOD NLL for Subspace, SWAG, Laplace, uncorrected random perturbations, and corrected P\&C. \textbf{(B--C)} Mean hidden change $\|\Delta h\|$ and output change $\|\Delta y\|$ on ID and Far-OOD data. Correction leaves the hidden perturbation large while suppressing the component of its output effect that transfers through the ID calibration geometry.}
    \label{fig:ant_teaser}
\end{figure}

A common strategy in post-hoc uncertainty estimation is to place substantial structure on the source of model variation, for example by sampling an approximate posterior or restricting variation to an SGD-derived subspace. P\&C instead places its main structural constraint on the correction. The perturbations are random and low-dimensional; the least-squares repair enforces agreement on calibration data and lets disagreement emerge where that repair no longer transfers. Figure~\ref{fig:ant_teaser} previews this mechanism empirically: corrected P\&C achieves a favorable ID/OOD frontier while maintaining large hidden perturbations whose output effect is suppressed on ID data and re-emerges under shift.

We evaluate P\&C on probabilistic MuJoCo dynamics prediction and CIFAR-10 OOD detection, and scale the same construction to a pretrained 86.6M-parameter ViT-B/16 on ImageNet-1K. Across these settings, P\&C preserves ID behavior while exposing useful disagreement under shift, although arbitrary shift need not increase uncertainty. An additional DistilBERT experiment tests the same construction under semantic shift. In our matched Deep Ensemble comparisons, P\&C substantially reduces the cost of constructing and storing the ensemble. Member-dependent inference remains $O(M)$, although computation before the earliest perturbed layer can be shared.

This work makes three contributions. First, we show that useful post-hoc epistemic ensembles can be constructed from a completed checkpoint by pairing random low-dimensional hidden-layer perturbations with local affine correction on calibration data. Second, we derive an exact account of the post-correction residual, including its transfer geometry, a leverage-based preservation bound, a covariance interpretation of ensemble disagreement, and corrected sensitivity as the infinitesimal limit. Third, we provide empirical evidence supporting this mechanism across dynamics, vision, and text, including matched frozen-checkpoint comparisons, sensitivity analyses, and explicit cost accounting. Code is available at \url{https://github.com/EtherealEq/perturb-and-correct}.

%% file: related_work.tex
\section{Related Work}
\label{sec:related-work}

\paragraph{Ensembles and approximate posterior sampling.}
A standard approach to epistemic uncertainty estimation is to aggregate predictions across multiple plausible predictors. Deep Ensembles remain a strong empirical baseline, with diversity arising from independent training runs~\citep{lakshminarayanan2017simple}, while MC Dropout obtains ensemble-like predictions by sampling stochastic subnetworks at test time~\citep{gal2016dropout}. Other post-training approaches construct approximate distributions around a trained solution. The Laplace approximation uses local curvature~\citep{ritter2018scalable}, SWAG fits a Gaussian approximation to an SGD trajectory~\citep{maddox2019simple}, and Subnetwork Linearized Laplace places a full-covariance Gaussian over selected internal weights in a linearized model~\citep{daxberger2021bayesian}. P\&C also represents uncertainty through an ensemble of predictors, but differs by constructing members from one trained checkpoint using hidden perturbations followed by calibration-preserving correction.

\paragraph{Low-loss geometry and structured parameter-space uncertainty.}
A related line of work studies structured high-performing regions in neural network parameter space. SWAG exploits the geometry traced by SGD iterates, while Subspace Inference performs posterior inference in a low-dimensional subspace often derived from optimization trajectories~\citep{maddox2019simple,izmailov2020subspace}. These methods show that useful model variation can often be found in structured low-dimensional regions of parameter space. P\&C shares the goal of obtaining uncertainty from low-dimensional parameter variation, but places the main inductive bias on what happens after the perturbation is applied: the perturbation is sampled randomly and the following affine layer is repaired on calibration data. The correcting step reduces the uncertainty to what can't be predicted from calibration data.

\paragraph{Single-model and frozen-checkpoint uncertainty.}
Several methods seek useful uncertainty estimates without training a full independent ensemble. SNGP combines spectral normalization with a Gaussian-process output layer~\citep{liu2020simple}, while Epistemic Neural Networks represent joint predictions through explicit epistemic indexing~\citep{osband2023epistemic}; both require changes to the model or training procedure. BatchEnsemble and Masksembles similarly introduce parameter sharing or structured masks during training~\citep{wen2020batchensemble,durasov2021masksembles}. A different class of methods operates directly on a completed predictor, including MSP, Energy, Mahalanobis distance, and ReAct~\citep{hendrycks2016baseline,liu2020energy,lee2018simple,sun2021react}. SCOD is especially relevant here: it sketches the training-data Fisher and measures test-point sensitivity along weakly constrained weight directions~\citep{sharma2021sketching}. P\&C is also applied after training, but returns nonlinear predictors and an ensemble predictive distribution rather than a single uncertainty score.

\paragraph{Adversarial sensitivity and constrained variation.}
The motivation underlying P\&C is also related to the adversarial example literature. Adversarial attacks expose the fact that a perturbation can be small under one set of constraints while producing a large predictive effect under another~\citep{goodfellow2014explaining,fawzi2018adversarial}, and robust optimization makes this asymmetry explicit through constrained worst-case variation~\citep{madry2017towards}. P\&C uses a related constrained-variation perspective in parameter space. Random perturbations are allowed, while affine repair keeps their ID-visible effects small on calibration data.

%% file: algorithm.tex
\section{Perturb-and-Correct}
\label{sec:method}

P\&C constructs an ensemble from a single trained network by perturbing one or more hidden-layer parameters and refitting a compensating correction in the following affine layer on calibration data. We treat the resulting predictors as a uniformly weighted mixture and combine predictions as in \citet{lakshminarayanan2017simple}. Throughout, let
\[
f_\theta = g_L \circ \phi \circ g_{L-1} \circ \cdots \circ \phi \circ g_1,
\]
where each $g_\ell(h)=W_\ell h+b_\ell$ is affine and $\phi$ is any fixed nonlinearity.\footnote{We present P\&C for a fully connected network for notational simplicity; Section~\ref{sec:experiments} applies it to convolutional and transformer blocks, and Appendix~\ref{app:conv_details} gives the corresponding architecture-specific forms.} Let $\mathcal D_{\mathrm{cal}}=\{x_i\}_{i=1}^N$ denote an ID calibration set, which is drawn from the data used to fit the base network.

\paragraph{Single-layer P\&C.}
Fix a target hidden layer $\ell$, ensemble size $M$, perturbation scale $\sigma>0$, and perturbation rank $K$. Draw a random orthonormal basis $U_\ell\in\mathbb R^{d_\ell\times K}$ once and share it across members, where $d_\ell$ is the number of perturbed parameters. For each member $m=1,\ldots,M$, sample $z^{(m)}\sim\mathcal N(0,I_K)$ and form
\begin{equation*}
\Delta W_\ell^{(m)}=\sigma\,\operatorname{reshape}\cdot(U_\ell z^{(m)}),
\qquad
\widetilde W_\ell^{(m)}=W_\ell+\Delta W_\ell^{(m)}.
\end{equation*}

Applying $\widetilde W_\ell^{(m)}$ changes the post-activation at layer $\ell$, which would otherwise propagate through the rest of the network. P\&C compensates locally by replacing the next affine layer $(W_{\ell+1},b_{\ell+1})$ with a corrected pair $(\widetilde W_{\ell+1}^{(m)},\widetilde b_{\ell+1}^{(m)})$ fit on calibration activations. Let $h_\ell(x)$ and $\widetilde h_\ell^{(m)}(x)$ denote the original and perturbed post-activations, and let $S_m\subseteq\mathcal D_{\mathrm{cal}}$ denote the correction subset for member $m$. The correction solves
\begin{equation}
\label{eq:PnC-correction}
\begin{aligned}
\bigl(\widetilde W_{\ell+1}^{(m)},\widetilde b_{\ell+1}^{(m)}\bigr)
=\arg\min_{W,b}\;&\sum_{x\in S_m}
\Bigl\|W\widetilde h_\ell^{(m)}(x)+b-
\bigl(W_{\ell+1}h_\ell(x)+b_{\ell+1}\bigr)\Bigr\|_2^2 \\
&+\lambda\|W-W_{\ell+1}\|_F^2+\lambda\|b-b_{\ell+1}\|_2^2.
\end{aligned}
\end{equation}
The regression target is the base model's next-layer pre-activation rather than a ground-truth label. The ridge penalty keeps the repair close to the pretrained affine map while allowing the calibration data to determine how much compensation is required. Each member shares every remaining parameter with the base model, so the only learned change after perturbation is the local affine repair.

\paragraph{Multi-layer P\&C.}
When perturbing a sequence of layers $\ell_1<\cdots<\ell_r$, P\&C applies the perturb-and-correct step sequentially. After perturbing layer $\ell_j$, it fits the correction in layer $\ell_j+1$ using activations produced by all preceding perturb-and-correct operations before proceeding to $\ell_{j+1}$. This yields a chain of local repairs rather than a single end-of-network correction. We perturb hidden layers only and leave output heads unchanged.

\paragraph{Bootstrapped correction.}
The correction subsets $\{S_m\}_{m=1}^M$ need not be shared across members. P\&C can introduce a second source of ensemble diversity by drawing each $S_m$ independently with replacement from $\mathcal D_{\mathrm{cal}}$, so that different members solve Eq.~\eqref{eq:PnC-correction} on different empirical correction geometries. Changing these subsets can decorrelate correction failures and increase the effective rank of the ensemble covariance, as discussed in Appendix~\ref{subsec:bootstrap-theory}. At the same time, reducing the correction set can make the least-squares problem unstable near its interpolation threshold. We therefore treat bootstrap fraction as controlling both diversity and correction stability.

%% file: theory.tex
\section{Theory}
\label{sec:theory}

We analyze P\&C as a randomized post-hoc ensemble mechanism. The central object is the \emph{post-correction residual}: the change at a test point that remains after the perturbed representation has been repaired by refitting the following affine layer on calibration data. We show (i) that this residual admits an exact decomposition under the applied perturbation; (ii) that its magnitude is controlled by a leverage term measuring how well the perturbed test representation is covered by the correction design; and (iii) that the local corrected-sensitivity expression and a covariance kernel for ensemble disagreement follow from the same residual. Together, these results give a geometric account of when correction continues to suppress a perturbation and when its effect can re-emerge away from the calibration geometry.

\subsection{Setup and exact correction}
\label{subsec:theory-setup}

Fix a trained network and a hidden layer whose post-activation representation is $y(x)\in\mathbb R^d$. Write the following affine layer in augmented form as
\begin{equation}
z(x)=\Theta\overline y(x),\qquad
\overline y(x):=\begin{bmatrix}1\\y(x)\end{bmatrix}\in\mathbb R^{d+1},
\qquad \Theta:=[b,W].
\label{eq:augmented-affine}
\end{equation}
A hidden-parameter perturbation $v$ replaces $\overline y$ by $\overline y_v$. Let $S=\{x_i\}_{i=1}^B$ denote a correction subset, and let $\overline Y_S,\overline Y_{v,S}$ stack the original and perturbed augmented representations respectively in rows, giving
\begin{equation}
\widehat\Theta_S(v):=\arg\min_{\Theta'}
\bigl\|\overline Y_{v,S}\Theta'^\top-\overline Y_S\Theta^\top\bigr\|_F^2
+\lambda\|\Theta'-\Theta\|_F^2.
\label{eq:ls-correction}
\end{equation}
Define
\[
G_{v,S}:=\overline Y_{v,S}^\top\overline Y_{v,S}+\lambda I,\qquad
\Delta Y_{v,S}:=\overline Y_{v,S}-\overline Y_S,\qquad
\Delta\overline y_v(x):=\overline y_v(x)-\overline y(x),
\]
and the post-correction residual
\begin{equation}
r_S(x;v):=\widehat\Theta_S(v)\overline y_v(x)-\Theta\overline y(x),
\qquad \rho_S(x;v):=\|r_S(x;v)\|_2.
\label{eq:post-correction-residual}
\end{equation}

\begin{lemma}[Exact correction and residual identity]
\label{thm:ls-optimal}
If $G_{v,S}$ is nonsingular,
\begin{equation}
\widehat\Theta_S(v)=
\Theta-\Theta\Delta Y_{v,S}^\top\overline Y_{v,S}G_{v,S}^{-1}.
\label{eq:ridge-correction-closed-form}
\end{equation}
With
\begin{equation}
\alpha_{v,S}(x):=\overline Y_{v,S}G_{v,S}^{-1}\overline y_v(x)
\label{eq:alpha}
\end{equation}
the residual is exactly
\begin{equation}
r_S(x;v)=\Theta\Bigl[\Delta\overline y_v(x)-\Delta Y_{v,S}^\top\alpha_{v,S}(x)\Bigr].
\label{eq:residual-identity}
\end{equation}
\end{lemma}

Equation~\eqref{eq:residual-identity} makes the action of the correction explicit. The test-point representation changes by $\Delta\overline y_v(x)$, while the least-squares fit reconstructs part of that change from the perturbation effects observed on the correction set. What remains after applying the following affine layer is the post-correction residual. Importantly, this identity uses the actual nonlinear representation changes induced by the perturbation rather than a Taylor expansion in $v$, so it describes the implemented correction without assuming a fixed activation region or first-order accuracy. Appendix~\ref{app:theory-proofs} gives the proof and the pseudoinverse extension for rank-deficient unregularized designs.

\subsection{ID preservation and the transfer defect}
\label{subsec:id-preservation}

Write the rows of $\Delta Y_{v,S}$ as $\Delta\overline y_v(x_i)^\top$. The bracketed quantity in Eq.~\eqref{eq:residual-identity} is the \emph{transfer defect}
\begin{equation}
g_S(x;v):=\Delta\overline y_v(x)-\sum_{i=1}^B\alpha_{v,S,i}(x)\Delta\overline y_v(x_i),
\qquad r_S(x;v)=\Theta g_S(x;v).
\label{eq:transfer-defect}
\end{equation}
The transfer defect measures how much of the test-point representation change is not reconstructed from the changes observed on the correction set under the same least-squares geometry. When calibration examples predict the perturbation effect at $x$ well, $g_S(x;v)$ is small and the repair continues to suppress the perturbation. When this relationship fails to transfer, the residual can remain large. Thus P\&C does not require every OOD input to produce high disagreement: an OOD point can still have low disagreement when its perturbation effects remain predictable from calibration data.

Define the perturbed-design ridge leverage
\begin{equation}
h_{v,S}^\lambda(x):=\overline y_v(x)^\top G_{v,S}^{-1}\overline y_v(x),
\label{eq:ridge-leverage}
\end{equation}
and perturbation magnitudes $\Delta_S(v):=\|\Delta Y_{v,S}\|_F$ and $\delta_x(v):=\|\Delta\overline y_v(x)\|_2$.

\begin{theorem}[Post-correction residual bound]
\label{thm:id-upper-bound}
Under Lemma~\ref{thm:ls-optimal},
\begin{equation}
\rho_S(x;v)\le
\|\Theta\|_{\mathrm{op}}\Bigl[\delta_x(v)+\Delta_S(v)\sqrt{h_{v,S}^\lambda(x)}\Bigr].
\label{eq:id-upper-bound}
\end{equation}
Moreover,
\begin{equation}
\|\alpha_{v,S}(x)\|_2^2
=h_{v,S}^\lambda(x)-\lambda\|G_{v,S}^{-1}\overline y_v(x)\|_2^2
\le h_{v,S}^\lambda(x).
\label{eq:alpha-leverage}
\end{equation}
\end{theorem}

The appearance of leverage gives the bound its geometric interpretation. When the perturbed representation $\overline y_v(x)$ is well covered by the correction design, $h_{v,S}^\lambda(x)$ is controlled and so are the reconstruction weights $\alpha_{v,S}(x)$. If the perturbation also changes nearby calibration representations in a way that predicts the change at $x$, the post-correction residual remains small. As leverage grows or this transfer relationship breaks, the bound becomes correspondingly less restrictive. Unlike the local sensitivity analysis below, the bound is pointwise in the applied perturbation and contains no Taylor remainder.

\subsection{Corrected sensitivity as the local limit}
\label{subsec:local-corollary}

The exact transfer defect also recovers the corrected-sensitivity decomposition studied in the local analysis. Suppose the representation maps are differentiable at $v=0$. Let
\[
J_x:=\partial_v\overline y_v(x)|_{v=0},\qquad
w_S^\lambda(x):=\overline Y_S(\overline Y_S^\top\overline Y_S+\lambda I)^{-1}\overline y(x),
\]
and define the calibration-predicted Jacobian
\[
T_S^\lambda(x):=\sum_{i=1}^B w_{S,i}^\lambda(x)J_{x_i}.
\]

\begin{corollary}[Corrected sensitivity]
\label{cor:local-sensitivity}
The derivative $A_S(x):=\partial_v r_S(x;v)|_{v=0}$ satisfies
\begin{equation}
A_S(x)=\Theta\bigl(J_x-T_S^\lambda(x)\bigr).
\label{eq:local-sensitivity-corollary}
\end{equation}
For $v\sim\mathcal N(0,\sigma^2I)$, $\mathbb E\|A_S(x)v\|_2^2=\sigma^2\|A_S(x)\|_F^2$.
\end{corollary}

This expression gives the same local geometric picture: corrected sensitivity is small when calibration Jacobians predict the test-point Jacobian and grows where that prediction fails. Random directions therefore probe the part of local sensitivity left unresolved by the calibration geometry without requiring knowledge of the eventual test-time shift. At the perturbation scales used in our experiments, however, the derivative at zero is not quantitatively accurate for the full member change. We therefore use the exact residual for the applied perturbations and treat Eq.~\eqref{eq:local-sensitivity-corollary} as its infinitesimal limit.

\subsection{Random perturbations and ensemble disagreement}
\label{subsec:ood-variance}

Let $V$ denote member randomness, including perturbation coefficients and, when used, the bootstrap correction subset. For each input let $\mu_S(x):=\mathbb E_Vr_V(x)$ and $\widetilde r_V(x):=r_V(x)-\mu_S(x)$, and define
\begin{equation}
K_S(x,x'):=\mathbb E_V[\widetilde r_V(x)\widetilde r_V(x')^\top].
\label{eq:covariance-kernel}
\end{equation}

\begin{theorem}[Random-function covariance kernel]
\label{thm:ood-variance}
If $\mathbb E_V\|r_V(x)\|_2^2<\infty$, then $K_S$ is a positive-semidefinite matrix-valued kernel. Its trace $k_S(x,x'):=\operatorname{tr}K_S(x,x')$ is a positive-semidefinite scalar kernel, and the usual sample covariance of i.i.d. P\&C members is an unbiased estimator of $K_S$.
\end{theorem}

The kernel describes how unresolved correction failures vary across the input space. Its diagonal $k_S(x,x)$ is the residual variance exposed by P\&C at $x$, while off-diagonal values identify points whose output-visible transfer defects co-vary across perturbations. In particular,
\begin{equation}
\mathbb E_V\|r_V(x)\|_2^2=\|\mu_S(x)\|_2^2+\operatorname{tr}K_S(x,x),
\qquad
\mathbb E_{V,V'}\|r_V(x)-r_{V'}(x)\|_2^2=2\operatorname{tr}K_S(x,x),
\label{eq:pairwise-disagreement}
\end{equation}
for independent members $V,V'$. Pairwise ensemble disagreement is therefore exactly twice the diagonal mass of this covariance kernel. Appendix~\ref{app:proof-covariance-kernel} gives the proof, and Appendix~\ref{subsec:bootstrap-theory} discusses how changing correction subsets can affect covariance rank.

%% file: experiments.tex
\begin{figure}[t]
    \centering
    \includegraphics[width=\linewidth]{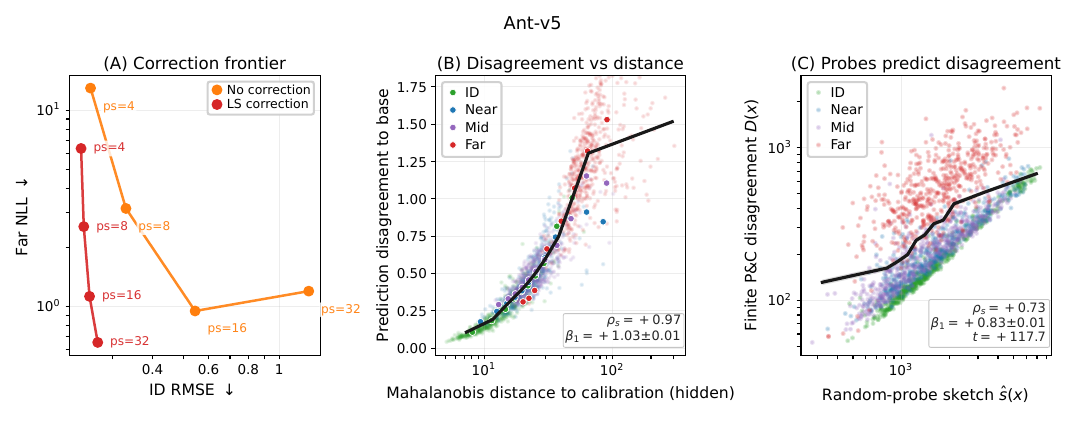}
    \caption{\textbf{Mechanism-level diagnostics for P\&C on Ant-v5.} \textbf{(A)} Least-squares correction improves the ID/OOD frontier: uncorrected random perturbations degrade ID behavior, while corrected P\&C achieves lower Far-OOD NLL at comparable ID error. \textbf{(B)} Prediction disagreement increases with Mahalanobis distance from the ID calibration activations. \textbf{(C)} The random-probe sketch from the infinitesimal corrected-sensitivity analysis remains correlated with the disagreement of the actual P\&C ensemble (Spearman $\rho_s=0.73$), although the local approximation is not quantitatively accurate at the experimental perturbation scale. Together, these diagnostics support the geometric picture in Section~\ref{sec:theory}: correction suppresses perturbation effects that are predictable from calibration data, while disagreement emerges where that relationship fails to transfer.}
    \label{fig:theory_bridge}
\end{figure}

\section{Experiments}
\label{sec:experiments}

\subsection{Experimental setup}

We evaluate Perturb-and-Correct (P\&C) in three main settings: probabilistic dynamics prediction under behavioral distribution shift in MuJoCo~\citep{todorov2012MuJoCo,towers2024gymnasium}, out-of-distribution (OOD) detection on CIFAR-10~\citep{krizhevsky2009learning} under the OpenOOD v1.5 protocol~\citep{zhang2023openood}, and ImageNet-1K OOD detection~\citep{deng2009imagenet} with a pretrained ViT-B/16~\citep{dosovitskiy2020image}. Across all settings, P\&C is constructed post hoc from a completed model, and all hyperparameters and post-hoc calibration choices are selected using in-distribution (ID) validation data only. We additionally report a DistilBERT Banking77 experiment in Appendix~\ref{app:distilbert} as a modality-transfer check.

MuJoCo uses 4,096 correction examples from the expert-transition training pool, CIFAR-10 uses 1,024 images from the training-fit split, and ImageNet uses 32,768 training images for correction with separate ID pools for selection and temperature fitting. Test and OOD examples never enter construction or selection. MuJoCo members are combined by moment matching their diagonal Gaussian predictions, with mean total predictive variance used for OOD ranking. For the classification experiments, we fit a scalar temperature on ID data, average member probabilities, and use predictive entropy as the P\&C OOD score. Full architecture, calibration, selection, and evaluation protocols are given in the appendices.

\subsection{Validating the Perturb-and-Correct mechanism}
\label{sec:mechanism-validation}

Before turning to aggregate benchmarks, we validate the mechanism described in Section~\ref{sec:theory}. The exact residual identity in Eq.~\eqref{eq:residual-identity} holds numerically to approximately $10^{-13}$ relative error in float64 on Ant, HalfCheetah, and Hopper across ID and shifted inputs. Propagating the corrected-layer residual through the frozen downstream Jacobian yields cosine alignment $0.97$--$1.00$ with the final output change. By contrast, the derivative-at-zero approximation is accurate only at perturbation norms roughly $30$--$300\times$ smaller than the operating range. We therefore use the exact residual to describe the applied perturbations and retain corrected sensitivity as its local limit. Appendix~\ref{app:mujoco-theory-validation} gives the validation protocol.

Figure~\ref{fig:theory_bridge}(A) compares random perturbations with and without least-squares correction on Ant-v5. Without correction, perturbations damage ID behavior; adding the repair shifts the frontier substantially, preserving ID performance without eliminating disagreement under Far-OOD shift. Figure~\ref{fig:theory_bridge}(B) shows the corresponding geometric pattern: P\&C disagreement increases with hidden Mahalanobis distance from the calibration activations. This relationship is not produced only by pooling ID and OOD regimes. On 40,000 points per environment, HalfCheetah and Hopper have pooled Spearman correlations $0.904$ and $0.749$, with within-tier correlations $0.816$--$0.860$ and $0.691$--$0.776$ respectively.

Finally, Figure~\ref{fig:theory_bridge}(C) connects the local corrected-sensitivity analysis to the actual ensemble. Although the derivative at zero is not quantitatively accurate at the applied perturbation scale, its random-probe sketch remains correlated with P\&C disagreement. Random-vs-Low diagnostics in Appendix~\ref{app:mujoco-sensitivity} further support random perturbation directions across Ant, HalfCheetah, and Hopper. Together, these results support the view that correction, rather than carefully chosen perturbation directions, is the main mechanism shaping useful disagreement.

\input{tables/gym_tables}

\subsection{MuJoCo: uncertainty under behavioral shift}
\label{sec:MuJoCo}

We evaluate multi-block P\&C on dynamics prediction across 11 MuJoCo environments. Models are evaluated on held-out expert transitions and progressively shifted policy data; Far transitions use uniform-random actions, with Near and Mid tiers available where provided by the benchmark. No shifted data are used for construction, selection, or calibration. Full per-environment results and hyperparameter details are in Appendix~\ref{app:MuJoCo}.

Table~\ref{tab:MuJoCo-cross-env} summarizes the comparison by cross-environment rank. P\&C is competitive on ID RMSE, with mean rank $3.18\pm2.40$ and five wins, and obtains the best mean rank on Far-OOD NLL ($1.55\pm1.21$), AUROC ($1.27\pm0.65$), and uncertainty--error Spearman correlation ($1.64\pm1.03$), with $8/11$, $9/11$, and $7/11$ wins respectively. The environment-level results are not uniformly favorable: ID error is mixed, HalfCheetah has notable variance, and Humanoid remains difficult. In a separate 27-seed graded-shift study, Far-AUROC reaches $0.998/0.987/0.980$ on Ant, HalfCheetah, and Hopper but falls to $0.343$ on Humanoid. Because this study uses a different run and aggregation protocol from the 10-seed validation-selected comparison, Appendix~\ref{app:MuJoCo} reports the two separately.

We also compare P\&C with SCOD from matched frozen checkpoints. P\&C has higher Far-AUROC in all 11 environments, with mean $0.920$ versus $0.512$ over the three matched seeds. A 27-seed layer-scope study gives Near/Mid/Far AUROC $0.764/0.815/0.904$ for first-block P\&C and $0.799/0.843/0.922$ for the multi-block construction. Multi-block correction is inexpensive for these small MLPs; the convolutional experiments instead select a single target block by ID-validation NLL to avoid compounding larger patchwise solves. Across the MuJoCo sensitivity sweeps, performance is broadly stable to rank and ridge, while perturbation scale exhibits an ID-stability boundary and correction-set size shows interpolation effects near the affine design dimension.

P\&C also retains the construction-cost advantage that motivates a post-hoc ensemble. For $M=50$, it is $46.8$--$61.7\times$ cheaper to construct than a matched Deep Ensemble on the three timed MuJoCo environments, adding only $1.5$--$1.8$ seconds beyond base training. This advantage concerns construction rather than inference: both methods still require $M$ member evaluations at test time.

\subsection{CIFAR-10: OOD detection}
\label{sec:cifar}

We next evaluate single-block P\&C on CIFAR-10 using a pre-activation ResNet-18~\citep{he2016identity} under the OpenOOD v1.5 protocol. Near OOD comprises CIFAR-100 and TinyImageNet, while Far OOD comprises MNIST, SVHN, Textures, and Places365; this protocol does not define a separate Mid tier. We compare against Mahalanobis, ReAct+Energy, LLLA, Epinet, SWAG, a Deep Ensemble, and matched SCOD. Target block, perturbation scale, and bootstrap fraction are selected by ID-validation NLL only.

\input{tables/cifar_tables}

Table~\ref{tab:cifar10_ood} shows that P\&C obtains the highest Near- and Far-OOD AUROC and the lowest corresponding FPR95 among the reported methods. Against matched SCOD, Near/Far AUROC improves from $89.69/92.56$ to $91.55/95.09$, while FPR95 falls from $39.40/21.41$ to $33.08/18.15$. The five-model Deep Ensemble retains the highest ID accuracy. As in MuJoCo, the main effect of P\&C is not to improve the base model's ID prediction, but to preserve that prediction while introducing model variation that becomes useful under shift.

We test whether these results depend on a narrow hyperparameter choice using a complete Cartesian product of six target blocks, three perturbation scales, and three bootstrap fractions over all three checkpoints, for 162 runs. Global ID-validation search and forward or reverse coordinate-wise selection recover the same optimum. Separate sweeps show nearly flat rank performance for $K\ge5$, ridge stability for $\lambda\ge10^{-3}$ but instability at zero, and a clear perturbation-scale ID-stability boundary (Appendix~\ref{app:cifar10-sweeps}). Representative early, middle, and late blocks remain viable.

The base ResNet-18 takes $8{,}010$ seconds to train. P\&C adds at most about $100$ seconds of post-training construction, while a matched 50-member Deep Ensemble would require approximately $400{,}500$ seconds by repeated base training. At inference, the two methods have essentially equal matched-$M$ latency (7.53 versus 7.26 ms/example). The computational advantage in this comparison is therefore the cost of constructing and storing the ensemble, not the online cost of evaluating it; Appendix~\ref{app:cifar-efficiency} gives the full timing and memory accounting.

\subsection{ImageNet-1K: large-scale transformer OOD detection}
\label{sec:imagenet-vit}

We next apply P\&C to the final feed-forward block of a pretrained 86.6M-parameter ViT-B/16 on ImageNet-1K. P\&C perturbs the first affine map of the final transformer FFN and refits the second affine map on final-block CLS representations. Because there is no later attention operation after this FFN, the correction can be fit on CLS rows alone; Appendix~\ref{app:transformer-block} gives the architectural details.

The primary configuration uses $K=20$ and $M=20$. Perturbation magnitude and ridge are selected using ID data only under a fixed 0.50 percentage-point top-1 preservation budget, selecting relative perturbation magnitude $r=2$ and $\lambda=1000$. The official 50,000-image ImageNet validation set is left untouched until final evaluation. The base checkpoint obtains $81.068\%$ top-1 accuracy, NLL $0.8482$, and ECE $0.0913$, while P\&C obtains $80.830\%$, $0.8045$, and $0.0542$, respectively.

\input{tables/imagenet_vit_tables}

Table~\ref{tab:imagenet_vit} shows that correction remains important at this scale. The same $r=2$ perturbations without correction obtain Near/Far AUROC $72.53/84.40$, while corrected P\&C obtains $76.49/87.89$, improvements of $3.96$ and $3.49$ points. P\&C also exceeds MSP on both aggregates and LLLA-Kron on Near OOD; against SCOD-linear it is stronger on Near and weaker on Far. A Mahalanobis detector on the final normalized CLS representation remains stronger for pure OOD ranking. Appendix~\ref{app:imagenet-vit} gives the preservation frontier, matched ablations, efficiency accounting, and additional mechanism diagnostics.

\paragraph{Limitations.}
These experiments do not establish universal effectiveness. P\&C depends on the calibration set covering the intended ID behavior, on the correction design remaining sufficiently well conditioned, and on perturbations staying within an ID-stable range. It can also be limited by the choice of target layer and the amount of useful downstream affine redundancy. More generally, arbitrary distribution shift need not increase disagreement, and dedicated representation-space scores can be stronger when the task is pure OOD ranking, as Mahalanobis is on ImageNet. Member-dependent inference remains $O(M)$, although late-layer constructions can reuse the shared network prefix. Broader evaluation across pretrained architectures, language models, and settings that require temporally coherent uncertainty remains important future work.

%% file: tables/gym_tables.tex
\begin{table}[t]
  \centering
  \small
  \caption{Cross-environment summary across 11 MuJoCo environments and 10 seeds. Entries report mean rank $\pm$ stddev and win counts; lower rank is better. Header arrows indicate the preferred direction of the underlying metric; superscript arrows indicate Wilcoxon-Holm comparisons against P\&C at $\alpha=0.05$: $\uparrow$ means P\&C significantly outperforms the baseline, and $\downarrow$ means the reverse.}
  \label{tab:MuJoCo-cross-env}
  \makebox[\textwidth][c]{%
  \begin{tabular}{lrrrrrrrr}
    \toprule
    Method & \multicolumn{2}{c}{ID RMSE ↓} & \multicolumn{2}{c}{Far NLL ↓} & \multicolumn{2}{c}{Far AUROC ↑} & \multicolumn{2}{c}{Far Spearman ↑} \\
     & mean rank & wins & mean rank & wins & mean rank & wins & mean rank & wins \\
    \midrule
    Deep Ensemble & 3.27 $\pm$ 1.56 & 2 / 11 & 3.00 $\pm$ 1.18 & 1 / 11 & 2.64 $\pm$ 1.36$^{\uparrow}$ & 1 / 11 & 2.27 $\pm$ 1.35 & 4 / 11 \\
    MC Dropout & 3.82 $\pm$ 1.40 & 0 / 11 & 3.45 $\pm$ 1.13 & 0 / 11 & 3.18 $\pm$ 1.72$^{\uparrow}$ & 1 / 11 & 4.18 $\pm$ 1.40$^{\uparrow}$ & 0 / 11 \\
    SWAG & 4.55 $\pm$ 1.44 & 0 / 11 & 4.00 $\pm$ 1.55 & 1 / 11 & 4.64 $\pm$ 1.21$^{\uparrow}$ & 0 / 11 & 4.91 $\pm$ 0.70$^{\uparrow}$ & 0 / 11 \\
    Laplace & 3.18 $\pm$ 1.47 & 1 / 11 & 3.09 $\pm$ 1.14 & 1 / 11 & 4.18 $\pm$ 0.87$^{\uparrow}$ & 0 / 11 & 2.91 $\pm$ 0.83$^{\uparrow}$ & 0 / 11 \\
    Subspace & \textbf{3.00 $\pm$ 1.73} & 3 / 11 & 5.91 $\pm$ 0.30$^{\uparrow}$ & 0 / 11 & 5.09 $\pm$ 0.83$^{\uparrow}$ & 0 / 11 & 5.09 $\pm$ 1.38$^{\uparrow}$ & 0 / 11 \\
    P\&C & 3.18 $\pm$ 2.40 & 5 / 11 & \textbf{1.55 $\pm$ 1.21} & 8 / 11 & \textbf{1.27 $\pm$ 0.65} & 9 / 11 & \textbf{1.64 $\pm$ 1.03} & 7 / 11 \\
    \bottomrule
  \end{tabular}
  }
\end{table}

%% file: tables/cifar_tables.tex
\begin{table}[t]
\centering
\small
\caption{CIFAR-10 OOD detection under OpenOOD v1.5 over three independently trained checkpoints (mean $\pm$ population standard deviation). Near OOD comprises CIFAR-100 and TinyImageNet; Far OOD comprises MNIST, SVHN, Textures, and Places365. SCOD uses the same checkpoints and a matched 1,024-example ID calibration pool. Bold and italic mark the best and second-best values.}
\label{tab:cifar10_ood}
\setlength{\tabcolsep}{4pt}
\begin{tabular}{lccccc}
\toprule
Method & Acc\% $\uparrow$ & Near AUROC $\uparrow$ & Near FPR95 $\downarrow$ & Far AUROC $\uparrow$ & Far FPR95 $\downarrow$ \\
\midrule
Mahalanobis & $95.74\pm0.14$ & $87.98\pm0.11$ & $66.30\pm0.88$ & $93.25\pm0.26$ & $38.30\pm5.44$ \\
ReAct+Energy & $\mathit{95.77\pm0.16}$ & $88.63\pm0.15$ & $54.39\pm1.28$ & $92.50\pm1.20$ & $29.78\pm5.26$ \\
LLLA & $\mathit{95.77\pm0.20}$ & $88.97\pm0.69$ & $54.12\pm5.70$ & $93.04\pm0.95$ & $28.44\pm6.72$ \\
Epinet & $95.76\pm0.18$ & $88.07\pm0.04$ & $63.37\pm1.40$ & $92.16\pm1.02$ & $35.10\pm5.57$ \\
SWAG & $95.37\pm0.05$ & $90.03\pm0.15$ & $44.71\pm0.79$ & $94.19\pm1.03$ & $22.09\pm4.91$ \\
SCOD & $95.74\pm0.14$ & $89.69\pm0.09$ & $\mathit{39.40\pm0.96}$ & $92.56\pm0.37$ & $21.41\pm0.87$ \\
P\&C & $95.59\pm0.20$ & $\mathbf{91.55\pm0.13}$ & $\mathbf{33.08\pm0.63}$ & $\mathbf{95.09\pm0.50}$ & $\mathbf{18.15\pm1.07}$ \\
Deep Ensemble & $\mathbf{96.56\pm0.04}$ & $\mathit{91.10\pm0.03}$ & $40.39\pm0.38$ & $\mathit{94.63\pm0.14}$ & $\mathit{19.49\pm0.26}$ \\
\bottomrule
\end{tabular}
\end{table}

%% file: tables/imagenet_vit_tables.tex
\begin{table}[t]
\centering
\small
\caption{ImageNet-1K OOD detection with a pretrained ViT-B/16. Near OOD comprises SSB-hard and NINCO; Far OOD comprises iNaturalist, Textures, and OpenImage-O. P\&C uses the primary 0.50 percentage-point ID-preservation budget, selecting $r=2$ and $\lambda=1000$ using ID data only. Uncorrected perturbation uses the identical perturbations without the affine repair. P\&C and Uncorrected are averaged over five post-hoc construction seeds.}
\label{tab:imagenet_vit}
\setlength{\tabcolsep}{7pt}
\begin{tabular}{lcc}
\toprule
Method & Near AUROC $\uparrow$ & Far AUROC $\uparrow$ \\
\midrule
MSP & 73.52 & 86.04 \\
Mahalanobis & \textbf{78.82} & \textbf{92.55} \\
LLLA-Kron & 72.95 & 87.42 \\
SCOD-linear & 75.15 & \textit{88.48} \\
Uncorrected perturb. & 72.53 & 84.40 \\
P\&C & \textit{76.49} & 87.89 \\
\bottomrule
\end{tabular}
\end{table}

%% file: appendix_theory.tex
\section{Proofs and additional theory for Section~\ref{sec:theory}}
\label{app:theory-proofs}

Throughout this appendix, abbreviate
\[
X:=\overline Y_S,\quad X_v:=\overline Y_{v,S},\quad \Delta X:=X_v-X,\quad
a:=\overline y(x),\quad a_v:=\overline y_v(x),\quad \Delta a:=a_v-a,
\]
and let $G_v:=X_v^\top X_v+\lambda I$.

\subsection{Proof of Lemma~\ref{thm:ls-optimal}}
\label{app:proof-ls-optimal}
\begin{proof}
Write $B:=\Theta'^\top$. The objective in Eq.~\eqref{eq:ls-correction} becomes
\[
\|X_vB-X\Theta^\top\|_F^2+\lambda\|B-\Theta^\top\|_F^2,
\]
with normal equations
\[
(X_v^\top X_v+\lambda I)B
=X_v^\top X\Theta^\top+\lambda\Theta^\top.
\]
Using $X=X_v-\Delta X$ gives
\[
B=\Theta^\top-G_v^{-1}X_v^\top\Delta X\,\Theta^\top,
\]
or, after transposing,
\[
\widehat\Theta_S(v)
=\Theta-\Theta\Delta X^\top X_vG_v^{-1},
\]
which is Eq.~\eqref{eq:ridge-correction-closed-form}. Multiplying by $a_v$ and subtracting $\Theta a$ gives Eq.~\eqref{eq:residual-identity}.
\end{proof}

\paragraph{Unregularized pseudoinverse extension.}
When $\lambda=0$ and $X_v$ is rank deficient, the minimum-norm solution is $\widehat\Theta_S(v)^\top=X_v^+X\Theta^\top$. Let $P_v:=X_v^+X_v$ and $\alpha_v^+(x):=(X_v^+)^\top a_v$. Then
\begin{equation}
r_S(x;v)=\Theta\Bigl[\Delta a-\Delta X^\top\alpha_v^+(x)-(I-P_v)a_v\Bigr].
\label{eq:pseudoinverse-residual}
\end{equation}
Thus a rank-deficient minimum-norm solve adds the explicit row-space projection defect $(I-P_v)a_v$.

\subsection{Proof of Theorem~\ref{thm:id-upper-bound}}
\label{app:proof-id-upper-bound}
\begin{proof}
From Eq.~\eqref{eq:residual-identity},
\begin{align*}
\rho_S(x;v)
&\le \|\Theta\|_{\mathrm{op}}
\bigl(\|\Delta a\|_2+\|\Delta X^\top\alpha_{v,S}(x)\|_2\bigr)\\
&\le \|\Theta\|_{\mathrm{op}}
\bigl(\delta_x(v)+\Delta_S(v)\|\alpha_{v,S}(x)\|_2\bigr).
\end{align*}
Since $X_v^\top X_v=G_v-\lambda I$,
\begin{align*}
\|\alpha_{v,S}(x)\|_2^2
&=a_v^\top G_v^{-1}X_v^\top X_vG_v^{-1}a_v\\
&=a_v^\top G_v^{-1}a_v-\lambda a_v^\top G_v^{-2}a_v\\
&=h_{v,S}^\lambda(x)-\lambda\|G_v^{-1}a_v\|_2^2\le h_{v,S}^\lambda(x).
\end{align*}
Substitution proves Eqs.~\eqref{eq:id-upper-bound} and~\eqref{eq:alpha-leverage}.
\end{proof}

\subsection{Transfer-defect identity}
\label{app:proof-transfer}
The $i$th row of $\Delta X$ is $\Delta\overline y_v(x_i)^\top$, so
\[
\Delta X^\top\alpha_{v,S}(x)=\sum_{i=1}^B\alpha_{v,S,i}(x)\Delta\overline y_v(x_i).
\]
Substitution into Eq.~\eqref{eq:residual-identity} gives Eq.~\eqref{eq:transfer-defect} and $r_S(x;v)=\Theta g_S(x;v)$.

\subsection{Covariance-kernel properties}
\label{app:proof-covariance-kernel}
For arbitrary inputs $x_1,\ldots,x_n$ and vectors $c_1,\ldots,c_n\in\mathbb R^q$,
\[
\sum_{i,j}c_i^\top K_S(x_i,x_j)c_j
=\mathbb E_V\left(\sum_i c_i^\top\widetilde r_V(x_i)\right)^2\ge0,
\]
so $K_S$ is a positive-semidefinite matrix-valued kernel. Its trace kernel is PSD by the same argument. For i.i.d. members, the usual sample cross-covariance is unbiased by the standard sample-covariance identity.

Writing $\mu_S(x)=\mathbb E_Vr_V(x)$ gives
\[
\mathbb E_V\|r_V(x)\|_2^2=\|\mu_S(x)\|_2^2+\operatorname{tr}K_S(x,x).
\]
For independent members $V,V'$ the centered cross term vanishes, yielding
\[
\mathbb E\|r_V(x)-r_{V'}(x)\|_2^2=2\operatorname{tr}K_S(x,x),
\]
as stated in Eq.~\eqref{eq:pairwise-disagreement}.

\subsection{Corrected sensitivity as the infinitesimal limit}
\label{app:proof-local-corollary}
At $v=0$, $\Delta a=\Delta X=0$ and
\[
\alpha_{0,S}(x)=X(X^\top X+\lambda I)^{-1}a=w_S^\lambda(x).
\]
Differentiating $g_S(x;v)=\Delta a(v)-\Delta X(v)^\top\alpha_{v,S}(x)$ at zero eliminates the derivative of $\alpha_{v,S}$ because it is multiplied by $\Delta X(0)=0$. Collecting perturbation coordinates gives
\[
A_S(x)=\Theta\bigl(J_x-T_S^\lambda(x)\bigr),
\]
which is Eq.~\eqref{eq:local-sensitivity-corollary}. For $v\sim\mathcal N(0,\sigma^2I)$,
$\mathbb E\|A_S(x)v\|_2^2=\sigma^2\|A_S(x)\|_F^2$.

\subsection{Bootstrapped correction and covariance alignment}
\label{subsec:bootstrap-theory}
For correction subset $S_j$, let $C_j(x):=K_{S_j}(x,x)\succeq0$ and define
\[
r_{\mathrm{eff}}(C):=\frac{\operatorname{tr}(C)^2}{\|C\|_F^2}.
\]

\begin{theorem}[Mixture rank versus covariance alignment]
\label{thm:bootstrap-rank}
Let $C_1,\ldots,C_M\in\mathbb R^{q\times q}$ be PSD matrices satisfying $\operatorname{tr}(C_j)=\tau$ and $\|C_j\|_F=\nu$. Define $\overline C=M^{-1}\sum_j C_j$ and
\begin{equation}
\overline\alpha:=\frac{1}{M(M-1)}\sum_{i\ne j}
\frac{\langle C_i,C_j\rangle_F}{\|C_i\|_F\|C_j\|_F}.
\label{eq:avg-alignment}
\end{equation}
Then
\begin{equation}
r_{\mathrm{eff}}(\overline C)
=\frac{\tau^2}{\nu^2}\frac{M}{1+(M-1)\overline\alpha},
\label{eq:bootstrap-rank-formula}
\end{equation}
which decreases monotonically with $\overline\alpha$.
\end{theorem}

\begin{proof}
Linearity gives $\operatorname{tr}(\overline C)=\tau$, while
\[
\|\overline C\|_F^2
=\frac{\nu^2}{M}\bigl(1+(M-1)\overline\alpha\bigr).
\]
Substitution into the effective-rank definition proves Eq.~\eqref{eq:bootstrap-rank-formula}.
\end{proof}

The equal-trace and equal-norm assumptions need not hold exactly empirically. The theorem is a structural explanation: changing correction geometry can increase covariance rank when transfer failures are weakly aligned, but bootstrapping does not guarantee such decorrelation and can become unstable near the interpolation threshold.

%% file: appendix_conv.tex
\section{Algorithmic details for convolutional and transformer networks}
\label{app:conv_details}

Section~\ref{sec:method} presents P\&C for a fully connected network. The CIFAR-10 experiments apply the same construction to a pre-activation ResNet-18, while the ImageNet experiment applies it to the feed-forward block of a Vision Transformer. In both cases the perturbation is followed by an affine map that can be refit by the same ridge-regression correction. The exact correction and post-correction residual analysis of Section~\ref{sec:theory} therefore carries over after reinterpreting the correction design for the architecture at hand.

\subsection{Convolutions as patchwise linear maps}
\label{app:conv-patch}

Let $X\in\mathbb R^{N\times H\times W\times C_{\rm in}}$ be a batch of feature maps and let a convolution have kernel side length $k$, stride $s$, and $C_{\rm out}$ output channels. Write the kernel in OIHW order as
\[
W_{\rm OIHW}\in\mathbb R^{C_{\rm out}\times C_{\rm in}\times k\times k}.
\]
Let
\[
P_s(X)\in\mathbb R^{(NH'W')\times(C_{\rm in}k^2)}
\]
be the unfolded patch matrix, with one receptive field per output position, and let $\operatorname{mat}(W_{\rm OIHW})\in\mathbb R^{(C_{\rm in}k^2)\times C_{\rm out}}$ flatten the kernel in the matching order. Then
\begin{equation}
\operatorname{flat}(\operatorname{Conv}(X,W))
=P_s(X)\operatorname{mat}(W_{\rm OIHW}).
\label{eq:conv-as-matmul}
\end{equation}
A broadcast bias is included by augmenting $P_s(X)$ with a column of ones, exactly as in the augmented representation $\overline y$ of Section~\ref{sec:theory}. Implementations storing kernels in HWIO order apply only a fixed permutation before and after the solve.

\subsection{Pre-activation residual blocks as perturb--correct units}
\label{app:conv-block}

We use a CIFAR-adapted pre-activation ResNet-18 with a $3\times3$ stride-1 stem, no max-pool, and four stages of two residual blocks with widths $64,128,256,512$. A basic block has the form
\begin{align}
y(x)&=\operatorname{Conv}_1(\operatorname{ReLU}(\operatorname{BN}_1(x));W_1),\\
u(x)&=\operatorname{ReLU}(\operatorname{BN}_2(y(x))),\\
z(x)&=\operatorname{Conv}_2(u(x);W_2),\\
\operatorname{Block}(x)&=z(x)+\operatorname{Shortcut}(x).
\end{align}
Batch-normalization layers use the running statistics of the trained base model and remain frozen during construction and inference.

For a selected block, P\&C perturbs $W_1$ and refits $W_2$ together with a new Conv$_2$ bias. The shortcut is unchanged, so matching the original Conv$_2$ output is equivalent to matching the residual branch contribution to the block output. All other network parameters remain frozen.

\subsection{Single-block P\&C}
\label{app:conv-singleblock}

Fix ensemble size $M$, perturbation scale $\sigma$, and rank $K$. Draw an orthonormal basis $U\in\mathbb R^{d_1\times K}$ in the flattened Conv$_1$ kernel space, where $d_1=k_1^2C_{{\rm in},1}C_{{\rm out},1}$. For member $m$,
\[
\Delta W_1^{(m)}=\sigma\,\operatorname{reshape}(Uz^{(m)}),\qquad
z^{(m)}\sim\mathcal N(0,I_K),
\]
and $\widetilde W_1^{(m)}=W_1+\Delta W_1^{(m)}$.

Let $U(x;W_1)$ denote the tensor entering Conv$_2$ after Conv$_1$, BN$_2$, and ReLU, and let $T_i^{\rm orig}$ be the original Conv$_2$ output for calibration image $x_i$. For member $m$, define the unfolded perturbed design
\[
Y_i^{(m)}:=P_{s_2}(U(x_i;\widetilde W_1^{(m)})),
\qquad
\overline Y_i^{(m)}=[\mathbf 1,Y_i^{(m)}].
\]
Writing $\Theta$ for the matrix that stacks the new bias and flattened Conv$_2$ kernel, and $\Theta_0$ for the original Conv$_2$ kernel augmented with zero bias, the correction is
\begin{equation}
\widehat\Theta^{(m)}
=\arg\min_{\Theta}
\sum_{i\in S_m}
\left\|\overline Y_i^{(m)}\Theta-T_i^{\rm orig,flat}\right\|_F^2
+\lambda\|\Theta-\Theta_0\|_F^2.
\label{eq:conv-correction}
\end{equation}
Thus the CIFAR implementation uses ridge centered on the original affine map.

Stacking rows over the member's correction subset gives $\overline Y_{S_m}^{(m)}$ and $T_{S_m}^{\rm orig}$. With
\[
H^{(m)}=\overline Y_{S_m}^{(m)\top}\overline Y_{S_m}^{(m)},
\]
the closed-form correction can be written as
\begin{equation}
\widehat\Theta^{(m)}
=\left(H^{(m)}+\lambda I\right)^{-1}
\left(\overline Y_{S_m}^{(m)\top}T_{S_m}^{\rm orig}+\lambda\Theta_0\right),
\label{eq:conv-ridge-soln}
\end{equation}
up to the transpose convention used to store output channels. This is the patchwise form of Eq.~\eqref{eq:ls-correction}.

The correction uses no class labels. Its targets are the base model's internal Conv$_2$ outputs on the calibration images. A learned Conv$_2$ bias is included even though the base convolution is bias-free, allowing the augmented constant feature to absorb perturbation-induced mean shifts.

\subsection{Chunked normal-equation accumulation}
\label{app:conv-chunked}

The bias-augmented correction dimension is
\[
D_M=1+k_2^2C_{{\rm in},2},
\]
which is at most 4,609 in the evaluated ResNet blocks. The full unfolded design, however, contains $|S_m|H'_2W'_2$ rows and need not fit comfortably in GPU memory. We therefore accumulate the sufficient statistics in chunks:
\begin{align}
H^{(m)}&\leftarrow H^{(m)}+\overline Y_c^{(m)\top}\overline Y_c^{(m)},\\
B^{(m)}&\leftarrow B^{(m)}+\overline Y_c^{(m)\top}T_c^{\rm orig}.
\label{eq:conv-chunk-update}
\end{align}
After all chunks are processed, Eq.~\eqref{eq:conv-ridge-soln} is solved once. Chunk size changes only memory usage and not the resulting normal equations. It is configured by block to fit the available GPU memory; the selected final-stage block uses chunk size 64 in the reported joint-grid and headline runs.

\subsection{Perturbation subspace}
\label{app:conv-subspace}

The random low-rank construction is identical to the fully connected case after flattening the Conv$_1$ kernel. We sample a Gaussian matrix in the $d_1$-dimensional kernel parameter space, orthonormalize its columns by QR factorization, and reshape sampled directions back to kernel form. The basis is shared across members and only the Gaussian coefficient vector varies. The selected CIFAR configuration uses $K=20$ and $M=50$.

\subsection{Target-block selection and multi-block variants}
\label{app:conv-multiblock}

The headline CIFAR-10 results use a single target block. We evaluate all six blocks in stages 2--4 in the joint block$\times$scale$\times$bootstrap search described in Appendix~\ref{app:cifar-joint-selection}; the first block of the final stage is selected by ID-validation NLL. The correction for this block has bias-augmented design dimension $1+9\cdot512=4{,}609$.

We also implement multi-block variants by applying the same local perturb-and-correct operation to multiple residual blocks in network order. Each selected block substitutes its member-specific perturbed Conv$_1$ and corrected Conv$_2$ parameters at inference, while all non-selected blocks retain the base parameters. These variants are used only for ablations and efficiency measurements in this paper; they are not used for the CIFAR headline comparison in Table~\ref{tab:cifar10_ood}.

\subsection{Transformer feed-forward blocks}
\label{app:transformer-block}

The ImageNet experiment applies P\&C to the final feed-forward block of a pretrained ViT-B/16. Writing the tokenwise feed-forward map as
\[
u=\phi(W_1h+b_1),\qquad z=W_2u+b_2,
\]
with GELU nonlinearity $\phi$, P\&C perturbs $W_1$ and refits $(W_2,b_2)$ to reproduce the base model's feed-forward output on ID calibration examples. The correction is therefore the same affine ridge problem as Eq.~\eqref{eq:PnC-correction}, with token representations playing the role of the design rows.

For the final transformer block, only the CLS row is needed in the correction fit. After this feed-forward block, the remaining operations before classification are tokenwise normalization followed by selection of the CLS representation and the classifier head; there is no later self-attention operation that can mix patch-token changes back into CLS. We therefore fit the reported ImageNet correction on final-block CLS representations. For an earlier transformer block, later attention can mix all token positions, so a corresponding correction would generally need to account for the relevant token rows rather than using CLS alone.

The ImageNet construction perturbs the first affine map of the final feed-forward block with a shared random rank-$20$ basis and uses ridge centered on the original second affine map. All attention parameters, LayerNorm parameters, and the classifier head remain fixed.

%% file: appendix_mujoco.tex
\section{MuJoCo Experimental Details}
\label{app:MuJoCo}

\subsection{Compute and implementation details}
\label{app:compute-details}

The MuJoCo runs were executed sequentially on a single machine with an Intel i7-12700KF CPU and an NVIDIA TITAN X (Pascal) with 12~GB of GPU memory under WSL2. The models use JAX and a $4\times200$ ReLU MLP with diagonal-Gaussian mean and variance heads. Unless otherwise stated, P\&C uses $K=20$ perturbation directions and $M=50$ members. The multi-block construction perturbs hidden layers 1 and 3 and corrects in the immediately following affine layers.

P\&C is strictly post-training. The correction pool contains 4,096 expert-training transitions, and validation uses a disjoint held-out slice of the expert data. Near and Mid shifts are drawn from the corresponding Minari policy tiers when available; Far data are collected using uniform-random actions. No shifted data are used for model construction, hyperparameter selection, temperature fitting, or other post-hoc calibration.

\subsection{Hyperparameter selection}
\label{app:gym-sweeps}

Every method in Table~\ref{tab:MuJoCo-cross-env} is selected independently per environment by ID-validation NLL. Final metrics are computed on disjoint ID and shifted evaluation data.

\paragraph{Shared protocol.}
\begin{description}
  \item[Architecture] $4\times200$ MLP with ReLU activations and a heteroscedastic Gaussian output head.
  \item[Training] $5{,}000$ optimizer steps, batch size $64$.
  \item[Ensemble size] Deep Ensemble uses $M=5$. P\&C uses $M=50$; MC Dropout, SWAG, Subspace, and Laplace use 100 predictive samples.
\end{description}

\paragraph{Candidate grids.}
\begin{description}
  \item[MC Dropout] Dropout probability in $\{0.05,0.10,0.20,0.30,0.50\}$; each value is trained separately.
  \item[SWAG] Sampling scale in $\{0.1,0.3,1.0,3.0,10.0\}$.
  \item[Subspace Inference] Temperature in $\{1,10,100,1000,10000\}$.
  \item[Laplace] Prior precision in $\{10,100,1000,10000,100000\}$.
  \item[P\&C] Perturbation size in $\{8,16,32\}$, bootstrap fraction in $\{0.05,0.10,0.20,0.30\}$, and ridge regularizer in $\{10^{-4},10^{-2}\}$, with $K=20$ and $M=50$ fixed.
\end{description}

All reported P\&C corrections use ridge centered on the pretrained affine map.

\subsection{Numerical validation of the residual analysis}
\label{app:mujoco-theory-validation}

The mechanism checks in Section~\ref{sec:mechanism-validation} use Ant-v5, HalfCheetah-v5, and Hopper-v5 at model seed 0 with float64 instrumentation. For the exact residual identity, we evaluate both perturb--correct interfaces, members $\{0,1,3,5\}$, $\lambda\in\{0,10^{-2},1,100\}$, and 1,024 points per available evaluation regime. Across correction, held-out ID, Near, Mid, and Far inputs, the direct post-correction residual and Eq.~\eqref{eq:residual-identity} agree to float64 precision: median relative errors are on the order of $10^{-13}$ and the observed maximum is at most $2.3\times10^{-11}$. The same stage-local identity also holds after an upstream perturb--correct stage has already modified the representations, with relative errors of approximately $10^{-13}$--$10^{-12}$ in the two-stage construction.

For the downstream bridge, we use the exact residual at the final corrected interface (hidden layer 3 corrected in layer 4). If $z_0$ is the base pre-activation entering the frozen output head $F_{\mathrm{down}}$, the actual output change is $\Delta y=F_{\mathrm{down}}(z_0+r)-F_{\mathrm{down}}(z_0)$. We compare this with $J_{\mathrm{down}}(z_0)r$, where $J_{\mathrm{down}}$ is the Jacobian of the frozen downstream map at $z_0$. Across regimes and tested scales, the cosine similarity is $1.00$ on Ant, $0.99$--$1.00$ on Hopper, and $0.97$--$1.00$ on HalfCheetah. This check concerns propagation of the \emph{exact} corrected-layer residual through the downstream network; it is distinct from linearizing the correction itself in the perturbation.

To test that latter approximation, we evaluate Corollary~\ref{cor:local-sensitivity} on the first perturb--correct interface over a logarithmic grid of perturbation norms, using three perturbation directions with $\lambda\in\{0,1\}$. Taking relative error at most $10\%$ as the quantitative-validity threshold, the derivative-at-zero approximation is accurate only for $\|\Delta W\|_F\lesssim0.04$--$0.16$ across the three environments, while the experimental operating range is $5$--$50$. Thus the applied perturbations are roughly $30$--$300\times$ beyond the range in which the infinitesimal approximation is quantitatively accurate.

\subsection{Sensitivity and direction diagnostics}
\label{app:mujoco-sensitivity}

The sensitivity study includes one-factor sweeps across all 11 environments and 27 seeds over perturbation scale, rank, bootstrap fraction, ridge, calibration-pool size, and layer scope. Each factor is varied around an environment-specific ID-selected anchor while all other settings are fixed. Rank is broadly stable through the tested central range; ridge has little predictive effect when correction is strongly overdetermined; and perturbation scale exhibits a clear ID-stability boundary. Near $n\approx p$, where $p=201$ includes the bias coordinate, the correction interpolates calibration targets while held-out residuals peak. A small ridge suppresses this double-descent peak by one to three orders of magnitude in Ant, HalfCheetah, and Hopper phase diagrams. The full-bootstrap curve uses a 10,000-example pool and is therefore interpreted as mechanism evidence rather than a directly matched one-factor comparison.

Across five seeds on Ant, HalfCheetah, and Hopper, Random-P\&C obtains higher Far-AUROC than Low-P\&C in every seed. On Ant, mean Far-AUROC increases from $0.47$ to $0.88$; on Hopper, from $0.81$ to $0.87$. Low-P\&C's ID-selected hidden perturbations shrink more strongly under shift and yield more aligned, lower-rank output changes, supporting random directions as a practical default.

\subsection{Graded shift and layer scope}
\input{tables/mujoco_pnc_tiers}

Across the 11-environment, 27-seed layer-scope study, macro-averaged Near/Mid/Far AUROC is $0.764/0.815/0.904$ for a first-block construction and $0.799/0.843/0.922$ for multi-block P\&C.

\subsection{SCOD protocol and results}
\label{app:scod-protocol}

SCOD is applied to the same frozen probabilistic MLP checkpoints and 4,096-point ID calibration sets. It constructs a randomized low-rank sketch of the dataset Fisher, with at most 100 eigenvectors and 604 sketch samples. Its hyperparameters are selected by ID-validation NLL before reading any OOD split. The comparison uses seeds $\{0,10,200\}$; the 27-seed P\&C aggregation yields the same per-environment win ordering.

\input{tables/mujoco_scod}

\subsection{Efficiency protocol and results}
\label{app:mujoco-efficiency-protocol}

Construction timing includes base-model training and all post-hoc work needed to materialize or sample $M=50$ predictors. The matched Deep Ensemble estimate uses 50 independently trained networks. P\&C compact storage records the base checkpoint and reconstructible perturbation/correction specification. In this MuJoCo implementation, both P\&C and a matched Deep Ensemble require $M$ full member forward evaluations, so the comparison does not claim an inference-time advantage.

\input{tables/mujoco_efficiency}

\subsection{Per-environment results}

Table~\ref{tab:MuJoCo-selected-hparams} gives the selected configurations. Table~\ref{tab:MuJoCo-per-env} reports ID RMSE, Far NLL, and Far AUROC by environment. Table~\ref{tab:MuJoCo-spearman} reports uncertainty--error Spearman correlations for the available tiers.

\input{tables/mujoco_selected_hparams}
\input{tables/mujoco_per_env_longtable}

\clearpage
\newpage

\input{tables/mujoco_spearman_longtable}

\clearpage
\newpage

%% file: tables/mujoco_pnc_tiers.tex
\begin{table}[t]
\centering
\small
\caption{P\&C OOD-detection AUROC across graded MuJoCo shift tiers in the separate 27-seed diagnostic aggregation. This study uses a different run and aggregation protocol from the 10-seed validation-selected headline benchmark, so its absolute AUROC values are not directly comparable to the per-environment headline table. Mid-tier results are available for these four environments; other environments either lack a distinct Mid dataset or were evaluated only at the shared Far tier.}
\label{tab:mujoco-pnc-tiers}
\begin{tabular}{lrrr}
\toprule
Environment & Near & Mid & Far \\
\midrule
Ant-v5 & 0.862 & 0.915 & 0.998 \\
HalfCheetah-v5 & 0.861 & 0.983 & 0.987 \\
Hopper-v5 & 0.896 & 0.947 & 0.980 \\
Humanoid-v5 & 0.570 & 0.522 & 0.343 \\
\bottomrule
\end{tabular}
\end{table}

%% file: tables/mujoco_scod.tex
\begin{table}[t]
\centering
\small
\caption{Matched frozen-checkpoint comparison with SCOD on MuJoCo. Far-AUROC is averaged over environments in the first column; representative environments are shown separately. SCOD and the seed-matched P\&C column each use seeds $\{0,10,200\}$.}
\label{tab:mujoco-scod}
\begin{tabular}{lccccc}
\toprule
Method & Mean & Ant & Hopper & Humanoid & Pusher \\
\midrule
SCOD & 0.512 & 0.110 & 0.028 & 0.267 & 0.995 \\
P\&C & \textbf{0.920} & \textbf{0.998} & \textbf{0.985} & \textbf{0.343} & \textbf{1.000} \\
\bottomrule
\end{tabular}
\end{table}

%% file: tables/mujoco_efficiency.tex
\begin{table}[t]
\centering
\small
\caption{Measured ensemble-construction time on an NVIDIA TITAN X (Pascal), with $M=50$. P\&C time includes one base-model training run and post-hoc member construction. The compact size stores the base checkpoint and reconstructible P\&C specification rather than 50 materialized members.}
\label{tab:mujoco-efficiency}
\begin{tabular}{lrrrrr}
\toprule
Environment & P\&C (s) & Deep Ens. (s) & Ratio & P\&C post-hoc (s) & Compact size (MB) \\
\midrule
Ant-v5 & 18.4 & 1131.9 & 61.7$\times$ & 1.75 & 0.76 \\
HalfCheetah-v5 & 27.0 & 1321.9 & 49.0$\times$ & 1.52 & 0.55 \\
Hopper-v5 & 27.3 & 1279.3 & 46.8$\times$ & 1.73 & 0.53 \\
\bottomrule
\end{tabular}
\end{table}

%% file: tables/mujoco_selected_hparams.tex
\begin{center}
  \scriptsize
  \captionof{table}{Selected hyperparameter configuration per (env, method), chosen by lowest mean validation NLL across the seeds used in the main-paper tables. \texttt{n}: ensemble members; \texttt{dr}: dropout rate; \texttt{sws}: SWAG schedule weight scale; \texttt{prior}: Laplace prior precision; \texttt{T}: subspace temperature; \texttt{lreg}, \texttt{bf}, \texttt{ps}: P\&C Tikhonov regulariser, per-member bootstrap fraction, and perturbation size.}
  \label{tab:MuJoCo-selected-hparams}
  \begin{tabular}{lllllll}
    \toprule
    Env & DE & MCD & SWAG & Lap. & Sub. & P\&C \\
    \midrule
    Ant-v5 & n=5 & dr=0.05 & sws=1.0 & prior=10000.0 & T=1.0 & lreg=0.0001, bf=0.1, ps=8.0 \\
    HalfCheetah-v5 & n=5 & dr=0.05 & sws=0.3 & prior=10000.0 & T=1.0 & lreg=0.0001, bf=0.3, ps=32.0 \\
    Hopper-v5 & n=5 & dr=0.05 & sws=0.1 & prior=10000.0 & T=1.0 & lreg=0.0001, bf=0.3, ps=32.0 \\
    Walker2d-v5 & n=5 & dr=0.05 & sws=0.1 & prior=10000.0 & T=1.0 & lreg=0.01, bf=0.3, ps=16.0 \\
    Swimmer-v5 & n=5 & dr=0.05 & sws=0.1 & prior=10000.0 & T=1.0 & lreg=0.0001, bf=0.05, ps=8.0 \\
    Humanoid-v5 & n=5 & default & sws=3.0 & prior=10000.0 & T=10.0 & lreg=0.01, bf=0.3, ps=32.0 \\
    HumanoidStandup-v5 & n=5 & default & sws=3.0 & prior=1000.0 & T=10.0 & lreg=0.01, bf=0.2, ps=8.0 \\
    Reacher-v5 & n=5 & dr=0.05 & sws=0.1 & prior=10000.0 & T=1.0 & lreg=0.0001, bf=0.05, ps=8.0 \\
    Pusher-v5 & n=5 & dr=0.05 & sws=1.0 & prior=10000.0 & T=10.0 & lreg=0.0001, bf=0.1, ps=8.0 \\
    InvertedPendulum-v5 & n=5 & dr=0.05 & sws=0.1 & prior=10000.0 & T=1.0 & lreg=0.0001, bf=0.05, ps=16.0 \\
    InvertedDoublePendulum-v5 & n=5 & dr=0.2 & sws=0.3 & prior=10000.0 & T=10.0 & lreg=0.0001, bf=0.05, ps=16.0 \\
    \bottomrule
  \end{tabular}
\end{center}

%% file: tables/mujoco_per_env_longtable.tex
\begingroup
\small
\begin{longtable}{lrrr}
\caption{Per-environment headline metrics on MuJoCo. Cells report mean $\pm$ stddev across seeds for the validation-NLL-selected configuration. Header arrows indicate the preferred direction. Within each environment, the best value(s) in each metric column are shown in \textbf{bold}.}
\label{tab:MuJoCo-per-env}\\
\toprule
Method & ID RMSE $\downarrow$ & Far NLL $\downarrow$ & Far AUROC $\uparrow$ \\
\midrule
\endfirsthead
\caption[]{Per-environment headline metrics on MuJoCo (continued)}\\
\toprule
Method & ID RMSE $\downarrow$ & Far NLL $\downarrow$ & Far AUROC $\uparrow$ \\
\midrule
\endhead
\midrule
\multicolumn{4}{r}{\emph{Continued on next page}} \\
\endfoot
\bottomrule
\endlastfoot
\multicolumn{4}{l}{\textit{Ant-v5} ($n=10$ seeds)} \\
    Deep Ensemble & 0.601 $\pm$ 0.022 & 14.361 $\pm$ 10.155 & 0.579 $\pm$ 0.075 \\
    MC Dropout & 0.571 $\pm$ 0.076 & 1374.124 $\pm$ 980.958 & 0.379 $\pm$ 0.120 \\
    SWAG & 0.674 $\pm$ 0.049 & \textbf{1.186 $\pm$ 0.358} & 0.476 $\pm$ 0.152 \\
    Laplace & 0.590 $\pm$ 0.093 & 6.438 $\pm$ 3.907 & 0.401 $\pm$ 0.167 \\
    Subspace & 0.552 $\pm$ 0.013 & 30.496 $\pm$ 24.740 & 0.302 $\pm$ 0.106 \\
    P\&C & \textbf{0.289 $\pm$ 0.151} & 1.544 $\pm$ 0.348 & \textbf{0.993 $\pm$ 0.003} \\
    \midrule
    \multicolumn{4}{l}{\textit{HalfCheetah-v5} ($n=10$ seeds)} \\
    Deep Ensemble & 1.323 $\pm$ 0.073 & 9.665 $\pm$ 5.843 & 0.969 $\pm$ 0.017 \\
    MC Dropout & 1.463 $\pm$ 0.103 & 6.087 $\pm$ 4.379 & 0.939 $\pm$ 0.026 \\
    SWAG & 1.493 $\pm$ 0.083 & 7.663 $\pm$ 5.000 & 0.964 $\pm$ 0.022 \\
    Laplace & \textbf{1.291 $\pm$ 0.088} & 5.568 $\pm$ 3.110 & 0.945 $\pm$ 0.024 \\
    Subspace & 1.641 $\pm$ 0.078 & 19.917 $\pm$ 34.424 & 0.955 $\pm$ 0.032 \\
    P\&C & 1.779 $\pm$ 0.800 & \textbf{3.735 $\pm$ 0.456} & \textbf{0.986 $\pm$ 0.010} \\
    \midrule
    \multicolumn{4}{l}{\textit{Hopper-v5} ($n=10$ seeds)} \\
    Deep Ensemble & \textbf{0.167 $\pm$ 0.010} & 2.637 $\pm$ 0.467 & 0.903 $\pm$ 0.019 \\
    MC Dropout & 0.210 $\pm$ 0.018 & 2.115 $\pm$ 0.683 & 0.624 $\pm$ 0.046 \\
    SWAG & 0.180 $\pm$ 0.009 & 4.579 $\pm$ 2.421 & 0.792 $\pm$ 0.048 \\
    Laplace & 0.175 $\pm$ 0.016 & 2.757 $\pm$ 0.823 & 0.641 $\pm$ 0.081 \\
    Subspace & 0.194 $\pm$ 0.008 & 5.443 $\pm$ 2.539 & 0.782 $\pm$ 0.055 \\
    P\&C & 0.219 $\pm$ 0.012 & \textbf{0.825 $\pm$ 0.080} & \textbf{0.979 $\pm$ 0.007} \\
    \midrule
    \multicolumn{4}{l}{\textit{Walker2d-v5} ($n=10$ seeds)} \\
    Deep Ensemble & 0.591 $\pm$ 0.023 & 9.450 $\pm$ 2.146 & 0.972 $\pm$ 0.005 \\
    MC Dropout & 0.566 $\pm$ 0.026 & 15.869 $\pm$ 3.836 & 0.945 $\pm$ 0.013 \\
    SWAG & 0.661 $\pm$ 0.016 & 28.842 $\pm$ 10.908 & 0.909 $\pm$ 0.030 \\
    Laplace & 0.579 $\pm$ 0.019 & 9.913 $\pm$ 2.415 & 0.897 $\pm$ 0.023 \\
    Subspace & 0.729 $\pm$ 0.016 & 36.106 $\pm$ 15.215 & 0.904 $\pm$ 0.030 \\
    P\&C & \textbf{0.551 $\pm$ 0.019} & \textbf{1.839 $\pm$ 0.029} & \textbf{0.991 $\pm$ 0.001} \\
    \midrule
    \multicolumn{4}{l}{\textit{Swimmer-v5} ($n=10$ seeds)} \\
    Deep Ensemble & 0.027 $\pm$ 0.012 & 3.498 $\pm$ 0.492 & \textbf{0.998 $\pm$ 0.005} \\
    MC Dropout & 0.058 $\pm$ 0.004 & 4.829 $\pm$ 0.645 & 0.992 $\pm$ 0.002 \\
    SWAG & 0.056 $\pm$ 0.086 & 38.232 $\pm$ 20.216 & 0.923 $\pm$ 0.076 \\
    Laplace & 0.035 $\pm$ 0.014 & 11.051 $\pm$ 2.425 & 0.988 $\pm$ 0.008 \\
    Subspace & \textbf{0.026 $\pm$ 0.006} & 76.561 $\pm$ 42.428 & 0.956 $\pm$ 0.024 \\
    P\&C & 0.051 $\pm$ 0.003 & \textbf{1.696 $\pm$ 0.439} & 0.990 $\pm$ 0.003 \\
    \midrule
    \multicolumn{4}{l}{\textit{Humanoid-v5} ($n=10$ seeds)} \\
    Deep Ensemble & 51.815 $\pm$ 6.334 & 4.234 $\pm$ 0.235 & 0.285 $\pm$ 0.053 \\
    MC Dropout & 44.184 $\pm$ 2.090 & 3.791 $\pm$ 0.385 & 0.288 $\pm$ 0.055 \\
    SWAG & 46.966 $\pm$ 8.429 & 3.745 $\pm$ 0.288 & 0.187 $\pm$ 0.027 \\
    Laplace & 64.620 $\pm$ 53.173 & 4.298 $\pm$ 1.239 & 0.251 $\pm$ 0.072 \\
    Subspace & 43.199 $\pm$ 2.519 & 4.377 $\pm$ 0.534 & 0.173 $\pm$ 0.024 \\
    P\&C & \textbf{27.215 $\pm$ 0.586} & \textbf{3.274 $\pm$ 0.090} & \textbf{0.837 $\pm$ 0.029} \\
    \midrule
    \multicolumn{4}{l}{\textit{HumanoidStandup-v5} ($n=10$ seeds)} \\
    Deep Ensemble & 33.783 $\pm$ 7.675 & 7.065 $\pm$ 0.546 & 0.543 $\pm$ 0.113 \\
    MC Dropout & 23.264 $\pm$ 3.640 & 6.258 $\pm$ 1.261 & 0.726 $\pm$ 0.072 \\
    SWAG & 23.211 $\pm$ 3.301 & 5.836 $\pm$ 0.638 & 0.575 $\pm$ 0.143 \\
    Laplace & 39.908 $\pm$ 26.810 & \textbf{5.614 $\pm$ 0.871} & 0.650 $\pm$ 0.122 \\
    Subspace & \textbf{21.107 $\pm$ 1.290} & 25.971 $\pm$ 22.086 & 0.577 $\pm$ 0.135 \\
    P\&C & 25.13 $\pm$ 5.226 & 5.757 $\pm$ 0.153 & \textbf{0.977 $\pm$ 0.017} \\
    \midrule
    \multicolumn{4}{l}{\textit{Reacher-v5} ($n=10$ seeds)} \\
    Deep Ensemble & \textbf{0.087 $\pm$ 0.009} & 2.587 $\pm$ 0.423 & 0.971 $\pm$ 0.004 \\
    MC Dropout & 0.111 $\pm$ 0.009 & 7.940 $\pm$ 0.930 & 0.929 $\pm$ 0.004 \\
    SWAG & 0.089 $\pm$ 0.011 & 190.001 $\pm$ 53.046 & 0.816 $\pm$ 0.055 \\
    Laplace & 0.088 $\pm$ 0.011 & 1.974 $\pm$ 0.404 & 0.926 $\pm$ 0.008 \\
    Subspace & 0.090 $\pm$ 0.009 & 1015.748 $\pm$ 653.260 & 0.771 $\pm$ 0.058 \\
    P\&C & 0.120 $\pm$ 0.021 & \textbf{0.917 $\pm$ 0.106} & \textbf{0.983 $\pm$ 0.003} \\
    \midrule
    \multicolumn{4}{l}{\textit{Pusher-v5} ($n=10$ seeds)} \\
    Deep Ensemble & 0.054 $\pm$ 0.009 & 5.899 $\pm$ 1.033 & 0.997 $\pm$ 0.006 \\
    MC Dropout & 0.057 $\pm$ 0.001 & 20.634 $\pm$ 4.510 & 0.972 $\pm$ 0.005 \\
    SWAG & 0.094 $\pm$ 0.051 & 28.527 $\pm$ 18.351 & 0.625 $\pm$ 0.309 \\
    Laplace & 0.054 $\pm$ 0.009 & 17.909 $\pm$ 2.376 & 0.900 $\pm$ 0.076 \\
    Subspace & \textbf{0.053 $\pm$ 0.002} & 2268.085 $\pm$ 1140.032 & 0.253 $\pm$ 0.091 \\
    P\&C & 0.054 $\pm$ 0.002 & \textbf{2.012 $\pm$ 0.227} & \textbf{1.000 $\pm$ 0.000} \\
    \midrule
    \multicolumn{4}{l}{\textit{InvertedPendulum-v5} ($n=10$ seeds)} \\
    Deep Ensemble & 0.017 $\pm$ 0.008 & -0.059 $\pm$ 0.811 & 0.832 $\pm$ 0.096 \\
    MC Dropout & 0.010 $\pm$ 0.002 & -0.508 $\pm$ 0.452 & 0.934 $\pm$ 0.010 \\
    SWAG & 0.052 $\pm$ 0.048 & 94.165 $\pm$ 65.723 & 0.488 $\pm$ 0.061 \\
    Laplace & 0.014 $\pm$ 0.014 & 1.178 $\pm$ 1.002 & 0.867 $\pm$ 0.148 \\
    Subspace & 0.015 $\pm$ 0.004 & 316.721 $\pm$ 181.217 & 0.559 $\pm$ 0.086 \\
    P\&C & \textbf{0.006 $\pm$ 0.002} & \textbf{-1.334 $\pm$ 0.447} & \textbf{0.956 $\pm$ 0.006} \\
    \midrule
    \multicolumn{4}{l}{\textit{InvertedDoublePendulum-v5} ($n=10$ seeds)} \\
    Deep Ensemble & 0.028 $\pm$ 0.013 & \textbf{4.230 $\pm$ 1.265} & 0.994 $\pm$ 0.004 \\
    MC Dropout & 0.030 $\pm$ 0.006 & 108.931 $\pm$ 75.299 & \textbf{0.997 $\pm$ 0.001} \\
    SWAG & 0.095 $\pm$ 0.057 & 124.220 $\pm$ 54.583 & 0.398 $\pm$ 0.289 \\
    Laplace & 0.023 $\pm$ 0.011 & 18.250 $\pm$ 4.784 & 0.951 $\pm$ 0.092 \\
    Subspace & 0.025 $\pm$ 0.005 & 2933.610 $\pm$ 2712.973 & 0.650 $\pm$ 0.215 \\
    P\&C & \textbf{0.020 $\pm$ 0.004} & 4.942 $\pm$ 0.542 & 0.995 $\pm$ 0.001 \\
\end{longtable}
\endgroup

%% file: tables/mujoco_spearman_longtable.tex
\begingroup
\small
\begin{longtable}{l@{\hskip 1.5em}rrrr}
\caption{Spearman rank correlation between predicted variance and squared error, computed per seed and reported as mean $\pm$ stddev across seeds. Header arrows indicate that higher is better. Within each environment, the best value(s) in each split column are shown in \textbf{bold}. Negative correlations are italicized to indicate anti-informative uncertainty signals.}
\label{tab:MuJoCo-spearman}\\
\toprule
Method & ID $\uparrow$ & Near $\uparrow$ & Mid $\uparrow$ & Far $\uparrow$ \\
\midrule
\endfirsthead
\caption[]{Spearman rank correlation between predicted variance and squared error (continued)}\\
\toprule
Method & ID $\uparrow$ & Near $\uparrow$ & Mid $\uparrow$ & Far $\uparrow$ \\
\midrule
\endhead
\midrule
\multicolumn{5}{r}{\emph{Continued on next page}} \\
\endfoot
\bottomrule
\endlastfoot
\multicolumn{5}{l}{\textit{Ant-v5}} \\
    Deep Ensemble & $0.32 \pm 0.04$ & $0.23 \pm 0.04$ & $0.08 \pm 0.05$ & $0.09 \pm 0.04$ \\
    MC Dropout & $0.38 \pm 0.15$ & $0.23 \pm 0.13$ & \emph{$-0.02 \pm 0.11$} & $0.06 \pm 0.07$ \\
    SWAG & $0.46 \pm 0.11$ & $0.25 \pm 0.07$ & \emph{$-0.08 \pm 0.06$} & $0.07 \pm 0.06$ \\
    Laplace & $0.44 \pm 0.09$ & $0.28 \pm 0.08$ & $0.04 \pm 0.11$ & $0.20 \pm 0.12$ \\
    Subspace & $0.30 \pm 0.04$ & $0.21 \pm 0.06$ & $0.01 \pm 0.05$ & $0.22 \pm 0.09$ \\
    P\&C & \textbf{$0.72 \pm 0.05$} & \textbf{$0.53 \pm 0.09$} & \textbf{$0.59 \pm 0.08$} & \textbf{$0.31 \pm 0.13$} \\
    \midrule
    \multicolumn{5}{l}{\textit{HalfCheetah-v5}} \\
    Deep Ensemble & $0.39 \pm 0.04$ & $0.61 \pm 0.06$ & \emph{$-0.09 \pm 0.09$} & $0.10 \pm 0.04$ \\
    MC Dropout & $0.39 \pm 0.08$ & $0.56 \pm 0.09$ & \emph{$-0.12 \pm 0.14$} & $0.08 \pm 0.05$ \\
    SWAG & $0.40 \pm 0.05$ & $0.42 \pm 0.10$ & \emph{$-0.09 \pm 0.15$} & $0.11 \pm 0.05$ \\
    Laplace & \textbf{$0.47 \pm 0.04$} & $0.57 \pm 0.09$ & \emph{$-0.12 \pm 0.19$} & $0.16 \pm 0.07$ \\
    Subspace & $0.37 \pm 0.06$ & $0.33 \pm 0.10$ & \emph{$-0.07 \pm 0.13$} & $0.13 \pm 0.05$ \\
    P\&C & $0.44 \pm 0.08$ & \textbf{$0.78 \pm 0.05$} & \textbf{$0.69 \pm 0.14$} & \textbf{$0.85 \pm 0.09$} \\
    \midrule
    \multicolumn{5}{l}{\textit{Hopper-v5}} \\
    Deep Ensemble & $0.65 \pm 0.02$ & \textbf{$0.64 \pm 0.02$} & $0.59 \pm 0.04$ & $0.52 \pm 0.05$ \\
    MC Dropout & $0.61 \pm 0.03$ & $0.35 \pm 0.06$ & $0.37 \pm 0.10$ & $0.55 \pm 0.07$ \\
    SWAG & \textbf{$0.67 \pm 0.03$} & $0.61 \pm 0.06$ & $0.05 \pm 0.18$ & $0.24 \pm 0.13$ \\
    Laplace & $0.66 \pm 0.06$ & $0.60 \pm 0.05$ & $0.49 \pm 0.09$ & $0.38 \pm 0.11$ \\
    Subspace & $0.64 \pm 0.03$ & $0.60 \pm 0.06$ & $0.09 \pm 0.19$ & $0.23 \pm 0.12$ \\
    P\&C & $0.49 \pm 0.07$ & $0.61 \pm 0.05$ & \textbf{$0.60 \pm 0.07$} & \textbf{$0.73 \pm 0.07$} \\
    \midrule
    \multicolumn{5}{l}{\textit{Walker2d-v5}} \\
    Deep Ensemble & $0.46 \pm 0.01$ & $0.45 \pm 0.03$ & $0.33 \pm 0.04$ & $0.26 \pm 0.03$ \\
    MC Dropout & \textbf{$0.51 \pm 0.03$} & $0.45 \pm 0.05$ & $0.35 \pm 0.08$ & $0.09 \pm 0.08$ \\
    SWAG & $0.47 \pm 0.02$ & $0.30 \pm 0.08$ & $0.34 \pm 0.05$ & $0.12 \pm 0.07$ \\
    Laplace & \textbf{$0.51 \pm 0.03$} & $0.39 \pm 0.05$ & $0.35 \pm 0.07$ & $0.20 \pm 0.08$ \\
    Subspace & $0.43 \pm 0.03$ & $0.26 \pm 0.07$ & $0.33 \pm 0.04$ & $0.09 \pm 0.05$ \\
    P\&C & \textbf{$0.51 \pm 0.02$} & \textbf{$0.50 \pm 0.05$} & \textbf{$0.36 \pm 0.02$} & \textbf{$0.36 \pm 0.04$} \\
    \midrule
    \multicolumn{5}{l}{\textit{Swimmer-v5}} \\
    Deep Ensemble & $0.53 \pm 0.14$ & $0.76 \pm 0.05$ & -- & $0.48 \pm 0.05$ \\
    MC Dropout & $0.46 \pm 0.09$ & $0.45 \pm 0.08$ & -- & $0.26 \pm 0.12$ \\
    SWAG & $0.59 \pm 0.16$ & \emph{$-0.19 \pm 0.23$} & -- & \emph{$-0.33 \pm 0.06$} \\
    Laplace & $0.41 \pm 0.25$ & $0.39 \pm 0.13$ & -- & \emph{$-0.21 \pm 0.08$} \\
    Subspace & $0.52 \pm 0.06$ & \emph{$-0.20 \pm 0.09$} & -- & \emph{$-0.29 \pm 0.07$} \\
    P\&C & \textbf{$0.65 \pm 0.09$} & \textbf{$0.80 \pm 0.04$} & -- & \textbf{$0.64 \pm 0.05$} \\
    \midrule
    \multicolumn{5}{l}{\textit{Humanoid-v5}} \\
    Deep Ensemble & $0.91 \pm 0.01$ & $0.64 \pm 0.03$ & \textbf{$0.42 \pm 0.10$} & \textbf{$0.76 \pm 0.03$} \\
    MC Dropout & $0.86 \pm 0.03$ & $0.56 \pm 0.06$ & $0.22 \pm 0.05$ & $0.62 \pm 0.04$ \\
    SWAG & $0.88 \pm 0.04$ & $0.57 \pm 0.07$ & $0.28 \pm 0.15$ & $0.69 \pm 0.07$ \\
    Laplace & \textbf{$0.92 \pm 0.02$} & \textbf{$0.65 \pm 0.11$} & $0.41 \pm 0.24$ & \textbf{$0.76 \pm 0.07$} \\
    Subspace & $0.84 \pm 0.05$ & $0.51 \pm 0.07$ & $0.17 \pm 0.08$ & $0.64 \pm 0.04$ \\
    P\&C & $0.83 \pm 0.02$ & \textbf{$0.65 \pm 0.03$} & \textbf{$0.42 \pm 0.05$} & $0.71 \pm 0.03$ \\
    \midrule
    \multicolumn{5}{l}{\textit{HumanoidStandup-v5}} \\
    Deep Ensemble & $0.65 \pm 0.18$ & $0.71 \pm 0.12$ & $0.84 \pm 0.08$ & $0.73 \pm 0.11$ \\
    MC Dropout & $0.18 \pm 0.14$ & $0.68 \pm 0.16$ & $0.82 \pm 0.09$ & $0.67 \pm 0.12$ \\
    SWAG & $0.30 \pm 0.17$ & $0.60 \pm 0.20$ & $0.78 \pm 0.12$ & $0.55 \pm 0.12$ \\
    Laplace & $0.61 \pm 0.26$ & $0.79 \pm 0.17$ & $0.90 \pm 0.07$ & $0.79 \pm 0.14$ \\
    Subspace & $0.18 \pm 0.11$ & $0.53 \pm 0.22$ & $0.74 \pm 0.11$ & $0.51 \pm 0.10$ \\
    P\&C & \textbf{$0.81 \pm 0.07$} & \textbf{$0.99 \pm 0.01$} & \textbf{$0.99 \pm 0.01$} & \textbf{$0.99 \pm 0.01$} \\
    \midrule
    \multicolumn{5}{l}{\textit{Reacher-v5}} \\
    Deep Ensemble & $0.60 \pm 0.08$ & $0.48 \pm 0.08$ & -- & \textbf{$0.85 \pm 0.02$} \\
    MC Dropout & $0.63 \pm 0.05$ & $0.55 \pm 0.07$ & -- & $0.73 \pm 0.03$ \\
    SWAG & $0.59 \pm 0.06$ & $0.39 \pm 0.12$ & -- & $0.43 \pm 0.20$ \\
    Laplace & $0.58 \pm 0.13$ & $0.43 \pm 0.11$ & -- & $0.79 \pm 0.02$ \\
    Subspace & $0.58 \pm 0.04$ & $0.46 \pm 0.06$ & -- & $0.34 \pm 0.20$ \\
    P\&C & \textbf{$0.69 \pm 0.06$} & \textbf{$0.63 \pm 0.08$} & -- & $0.70 \pm 0.04$ \\
    \midrule
    \multicolumn{5}{l}{\textit{Pusher-v5}} \\
    Deep Ensemble & $0.59 \pm 0.06$ & $0.74 \pm 0.08$ & -- & \textbf{$0.78 \pm 0.02$} \\
    MC Dropout & \textbf{$0.65 \pm 0.03$} & $0.70 \pm 0.02$ & -- & $0.45 \pm 0.05$ \\
    SWAG & $0.33 \pm 0.15$ & $0.14 \pm 0.33$ & -- & $0.24 \pm 0.35$ \\
    Laplace & $0.62 \pm 0.05$ & $0.58 \pm 0.19$ & -- & $0.62 \pm 0.09$ \\
    Subspace & $0.58 \pm 0.03$ & $0.31 \pm 0.09$ & -- & \emph{$-0.47 \pm 0.09$} \\
    P\&C & $0.61 \pm 0.05$ & \textbf{$0.85 \pm 0.04$} & -- & $0.77 \pm 0.02$ \\
    \midrule
    \multicolumn{5}{l}{\textit{InvertedPendulum-v5}} \\
    Deep Ensemble & \textbf{$0.76 \pm 0.14$} & $0.76 \pm 0.10$ & -- & $0.69 \pm 0.08$ \\
    MC Dropout & $0.50 \pm 0.10$ & $0.59 \pm 0.06$ & -- & $0.54 \pm 0.08$ \\
    SWAG & $0.61 \pm 0.20$ & \emph{$-0.24 \pm 0.22$} & -- & \emph{$-0.57 \pm 0.25$} \\
    Laplace & $0.46 \pm 0.22$ & $0.41 \pm 0.34$ & -- & $0.49 \pm 0.41$ \\
    Subspace & $0.46 \pm 0.09$ & \emph{$-0.01 \pm 0.13$} & -- & \emph{$-0.58 \pm 0.22$} \\
    P\&C & $0.41 \pm 0.18$ & \textbf{$0.90 \pm 0.02$} & -- & \textbf{$0.86 \pm 0.03$} \\
    \midrule
    \multicolumn{5}{l}{\textit{InvertedDoublePendulum-v5}} \\
    Deep Ensemble & $0.57 \pm 0.17$ & $0.44 \pm 0.12$ & -- & \textbf{$0.88 \pm 0.02$} \\
    MC Dropout & $0.35 \pm 0.10$ & $0.45 \pm 0.10$ & -- & $0.44 \pm 0.08$ \\
    SWAG & $0.31 \pm 0.45$ & $0.12 \pm 0.33$ & -- & \emph{$-0.52 \pm 0.16$} \\
    Laplace & $0.36 \pm 0.21$ & $0.27 \pm 0.18$ & -- & $0.68 \pm 0.17$ \\
    Subspace & $0.18 \pm 0.10$ & $0.10 \pm 0.13$ & -- & \emph{$-0.39 \pm 0.25$} \\
    P\&C & \textbf{$0.60 \pm 0.15$} & \textbf{$0.78 \pm 0.07$} & -- & $0.85 \pm 0.03$ \\
\end{longtable}
\endgroup

%% file: appendix_cifar.tex
\section{CIFAR-10 protocol, selection, and sensitivity}
\label{app:cifar10-sweeps}

\subsection{Data splits and selection protocol}
\label{app:cifar10-protocol}

CIFAR-10 contains 50,000 training and 10,000 test images. We reserve 10\% of the training set (5,000 images) as an ID validation split using a fixed seed and use the remaining 45,000 images for base-model training. Hyperparameters are selected by ID-validation NLL after post-hoc temperature scaling; no OOD data are used for construction, hyperparameter selection, or temperature fitting.

P\&C uses a 1,024-example correction pool drawn without replacement from the 45,000-image training-fit split. For a given checkpoint, the same base correction pool is reused across hyperparameter configurations. When bootstrapping is enabled, each member resamples within this pool. The affine correction uses only intermediate activations and the base model's pre-activation targets; class labels are not used by the correction solve. Validation, CIFAR-10 test, and OOD examples never enter the correction fit.

For classifier inference, one scalar temperature is fitted per configuration on the 5,000-example ID-validation split. The same temperature is applied to every member's logits before softmax, member probabilities are averaged, and predictive entropy is used as the P\&C OOD score. FPR95 is the false-positive rate at the first threshold reaching at least 95\% true-positive rate. Near and Far metrics are arithmetic means over the datasets in their respective OpenOOD groups.

The shared PreAct ResNet-18 training recipe uses SGD with Nesterov momentum 0.9, initial learning rate $0.1$ with cosine decay and a five-epoch warmup, weight decay $5\times10^{-4}$, batch size 128, 300 epochs, flip/crop plus cutout(8) augmentation, and no label smoothing.

\subsection{Hyperparameter grids}
\label{app:cifar10-grid}

Table~\ref{tab:cifar10-sweep-grid} lists the principal selection grids. Secondary settings not shown were fixed to their implementation or literature defaults.

\begin{table}[h]
\centering
\caption{CIFAR-10 hyperparameter grids. Bold marks the selected P\&C values. The P\&C target block, scale, and bootstrap fraction were evaluated jointly as a complete Cartesian product; rank and ridge were studied separately in one-factor sensitivity sweeps.}
\label{tab:cifar10-sweep-grid}
\small
\setlength{\tabcolsep}{4pt}
\begin{tabular}{lll}
\toprule
Method & Hyperparameter & Values \\
\midrule
MC Dropout & dropout rate & $\{0.1,\mathbf{0.2},0.3,0.5\}$ \\
LLLA & prior precision & $\{0.01,0.1,1,\mathbf{10},100,1000\}$ \\
SWAG & start epoch & $\{160,\mathbf{240}\}$ \\
Epinet & prior scale & $\{0.5,1,\mathbf{3}\}$ \\
Deep Ensemble & number of models & $5$ \\
SCOD & calibration size / sketch rank & $1024 / 10$ \\
\midrule
P\&C & target (stage, block) & $\{1,2,3\}\times\{0,1\}$; $\mathbf{(3,0)}$ \\
 & perturbation scale $\sigma$ & $\{\mathbf{25},50,100\}$ \\
 & bootstrap fraction & $\{\mathbf{0.05},0.10,0.20\}$ \\
 & perturbation rank $K$ & $20$ \\
 & ridge $\lambda$ & $10^{-3}$ \\
 & ensemble size / correction pool & $50 / 1024$ \\
\bottomrule
\end{tabular}
\end{table}

\subsection{Joint P\&C selection}
\label{app:cifar-joint-selection}

To test interactions among the three principal P\&C selection dimensions, we evaluate all six target blocks, three perturbation scales, and three bootstrap fractions: $6\times3\times3=54$ configurations per checkpoint and 162 runs across checkpoints/seeds $\{0,1,2\}$. All configurations use $K=20$, $M=50$, $\lambda=10^{-3}$, a 1,024-example correction pool, random directions, frozen batch-normalization statistics, and ridge centered on the original corrected convolution.

Selection minimizes mean temperature-scaled ID-validation NLL across the three checkpoints. The global optimum within this grid is stage 3, block 0 (the first block of the final ResNet stage), $\sigma=25$, and bootstrap fraction $0.05$, with validation NLL $0.1305\pm0.0049$. This configuration ranks first among all 54 combinations. Coordinate-wise selection in either forward or reverse order reaches the same configuration, so the selected result is not an artifact of the coordinate-wise search procedure. OOD datasets are loaded only after this configuration is fixed.

\subsection{One-factor sensitivity}
\label{app:cifar-sensitivity}

\input{tables/cifar_sensitivity}

The scale sweep shows the clearest stability boundary: performance improves through $\sigma=25$ but large perturbations cross the ID-preservation regime, with mean ID accuracy falling to roughly 35\% at $\sigma=50$ and 10\% at $\sigma=100$. Rank is much less sensitive: Near AUROC is essentially flat from $K=5$ through $K=40$. Ridge has a threshold effect rather than being irrelevant. The unregularized solve is singular or numerically unstable, $\lambda=10^{-4}$ is degraded, and $\lambda\in[10^{-3},1]$ produces nearly identical predictive performance.

Early, middle, and late target blocks all yield viable predictors at the anchor settings. Their mean accuracies span $94.76$--$95.59\%$, Near AUROC $89.34$--$91.55$, and Far AUROC $92.01$--$95.09$; the final-stage first block is selected because it has the lowest ID-validation NLL ($0.130$ versus $0.147$ and $0.151$ for the representative early and middle blocks).

Bootstrap fraction is conditioning-sensitive. In the completed joint grid at the selected block and scale, fractions $0.05$, $0.10$, and $0.20$ yield validation NLL $0.1305$, $0.1357$, and $0.1452$, respectively. A separate refined one-factor harness produces an instability at $0.10$ that does not reproduce in the joint grid, so we do not interpret that isolated collapse as a robust double-descent effect. A calibration-size sweep likewise contains a reproducible numerical/conditioning failure at 2,048 examples while neighboring sizes remain healthy, reinforcing that correction conditioning, rather than sample count alone, is an important practical constraint.

\subsection{Matched SCOD comparison}
\label{app:cifar-scod}

SCOD and P\&C use the same three PreAct ResNet-18 checkpoints and matched 1,024-example ID calibration pools. The SCOD construction uses sketch rank 10 and is selected without OOD data. Table~\ref{tab:cifar10_ood} compares SCOD with the same P\&C configuration selected above. SCOD obtains Near/Far AUROC $89.69/92.56$, while P\&C obtains $91.55/95.09$; corresponding FPR95 values are $39.40/21.41$ for SCOD and $33.08/18.15$ for P\&C.

\subsection{Efficiency, storage, and memory}
\label{app:cifar-efficiency}

The measured base-model training time is 8,010 seconds. P\&C's post-training construction was not instrumented as a standalone timer in the anchor run; subtracting the measured OpenOOD evaluation from the combined build-and-evaluate run gives a conservative upper bound of approximately 100 seconds. We therefore report this quantity explicitly as an upper bound rather than a direct measurement. Deep-Ensemble construction time in Table~\ref{tab:cifar-efficiency} is derived by multiplying the measured base-training time by the number of independently trained members.

\input{tables/cifar_efficiency}

At matched $M=50$, P\&C and the Deep Ensemble have essentially the same measured inference latency (7.53 versus 7.26 ms/example), as expected because both perform 50 member forward passes. P\&C therefore has no matched-member inference-time advantage. Its efficiency advantage relative to a 50-member Deep Ensemble is in avoiding 49 additional training runs and in persistent representation: the benchmark implementation reconstructs P\&C from one 42.7-MiB base checkpoint, while 50 independent checkpoints require approximately 2.13 GiB. The memory measurements show that this construction advantage does not imply uniformly lower peak GPU memory; P\&C peaks at 4.33 GiB in the measured run versus 4.07 GiB for the matched Deep Ensemble, while SWAG and LLLA have different storage and memory tradeoffs.

%% file: tables/cifar_sensitivity.tex
\begin{table}[t]
\centering
\small
\caption{CIFAR-10 P\&C sensitivity around the ID-selected configuration. Each entry aggregates checkpoints/seeds $\{0,1,2\}$. Scale, rank, ridge, and target block are one-factor sweeps; bootstrap values are from the joint block$\times$scale$\times$bootstrap grid at the selected block and scale.}
\label{tab:cifar-sensitivity}
\setlength{\tabcolsep}{3pt}
\begin{tabular}{l p{0.9in} p{0.8in} p{2.5in}}
\toprule
Factor & Sweep & Anchor & Main observation \\
\midrule
Scale $\sigma$ & 5 values, $6.25$--$100$ & $25$ & Near AUROC peaks at $91.55$; ID accuracy collapses for $\sigma\ge50$ \\
Rank $K$ & 5 values, $1$--$40$ & $20$ & Near AUROC is $91.40$--$91.55$ for $K\ge5$ \\
Ridge $\lambda$ & 6 values, $0$--$1$ & $10^{-3}$ & $\lambda=0$ is singular/unstable; $\lambda\ge10^{-3}$ is nearly flat \\
Target block & early, middle, late & late (s3b0) & val NLL $0.147/0.151/0.130$; all three remain viable \\
Bootstrap frac. & $.05,.10,.20$ & $.05$ & val NLL $0.1305/0.1357/0.1452$ at s3b0, $\sigma=25$ \\
\bottomrule
\end{tabular}
\end{table}

%% file: tables/cifar_efficiency.tex
\begin{table}[t]
\centering
\small
\caption{CIFAR-10 construction and matched-member inference accounting. Inference uses $5{,}000$ images, batch size 256, one synchronized warm run on seed 0. P\&C construction is a derived upper bound beyond one measured $8{,}010$-s base training run; Deep Ensemble training is derived as $50\times$ the same base-training time. SWAG and LLLA report measured post-training construction.}
\label{tab:cifar-efficiency}
\setlength{\tabcolsep}{5pt}
\begin{tabular}{lccc}
\toprule
Method & Training requirement & Post-training construction & Inference (ms/example) \\
\midrule
P\&C ($M=50$) & 1 base ($8{,}010$ s) & $\le 100$ s (upper bound) & 7.53 \\
Deep Ensemble ($M=50$) & 50 bases ($\approx400{,}500$ s) & -- & 7.26 \\
SWAG ($M=50$) & 1 trajectory fit & $55.8\pm5.1$ s & 7.33 \\
LLLA ($M=50$) & 1 base & $15.56\pm0.30$ s & 7.68 \\
\bottomrule
\end{tabular}
\end{table}

\begin{table}[t]
\centering
\small
\caption{CIFAR-10 storage and GPU-memory accounting. GPU resident and peak memory are measured for seed 0 with JAX preallocation disabled. Persistent P\&C storage is the $42.7$-MiB base checkpoint because members are reconstructed; materializing the 50 compact convolutional corrections would require approximately $450$ MiB (derived). Deep-Ensemble storage is derived as $50\times$ the measured base-checkpoint size.}
\label{tab:cifar-memory}
\setlength{\tabcolsep}{6pt}
\begin{tabular}{lccc}
\toprule
Method & Persistent representation (MiB) & Resident GPU (MiB) & Peak GPU (MiB) \\
\midrule
P\&C ($M=50$) & $42.7$ reconstructed ($\sim450$ materialized) & 1,063 & 4,332 \\
Deep Ensemble ($M=50$) & $2{,}133$ (derived) & 2,250 & 4,072 \\
SWAG ($M=50$) & $980$ & 1,071 & 5,360 \\
LLLA ($M=50$) & -- & 245 & 2,097 \\
\bottomrule
\end{tabular}
\end{table}

%% file: appendix_imagenet.tex
\section{ImageNet-1K ViT experiment}
\label{app:imagenet-vit}

We evaluate P\&C on a pretrained ViT-B/16 ImageNet-1K checkpoint with 86.6M parameters. The construction targets the final transformer feed-forward block: P\&C perturbs the first affine map of the FFN, applies the fixed GELU nonlinearity, and refits the second affine map. As described in Appendix~\ref{app:transformer-block}, the correction is fit on final-block CLS representations rather than patch-token rows.

We partition ImageNet training examples into disjoint pools for correction (32,768 images), ID selection (8,192), and temperature fitting (8,192). The official 50,000-image validation set is left untouched until final evaluation. P\&C uses perturbation rank $K=20$ and final ensemble size $M=20$. Member probabilities are temperature-scaled with a shared scalar temperature $T=0.700$ fit on the ID temperature pool, then averaged; predictive entropy is used for OOD ranking. Near OOD comprises SSB-hard and NINCO, while Far OOD comprises iNaturalist, Textures, and OpenImage-O.

\subsection{ID-preservation selection}
\label{app:imagenet-preservation}

The perturbation scale is selected by asking how large a perturbation the corrected model can support while preserving ID accuracy. The primary operating point uses a fixed 0.50 percentage-point top-1 preservation budget. For each candidate perturbation magnitude $r=\|\Delta W_1\|_F/\|W_1\|_F$, ridge strength is selected using ID data only, and preservation is checked using a one-sided 95\% lower confidence bound from a paired bootstrap over the ID-selection examples. The matched uncorrected ensemble is an ablation and does not enter selection.

This procedure selects $r=2.0$ and $\lambda=1000$ for the primary budget. The reported $\lambda$ corresponds to the summed least-squares objective in Eq.~\eqref{eq:PnC-correction}; with 32,768 CLS correction rows, its mean-objective equivalent is approximately $0.0305$. The corrected ensemble passes through $r=2.0$ and fails the primary preservation criterion by $r=2.125$. For sensitivity, 0.25- and 1.00-percentage-point budgets select $r=1.25$ and $r=3.0$, respectively, with the same selected ridge. These operating points are fixed before the follow-up OOD evaluation. The preservation-budget protocol itself was developed after an initial feasibility study that had inspected OOD behavior at several perturbation scales; we therefore treat the follow-up as an ID-selected diagnostic rather than a preregistered OOD study.

On the official ImageNet validation set, the base checkpoint obtains 81.068\% top-1 accuracy, NLL 0.8482, and ECE 0.0913. The primary P\&C ensemble obtains 80.830\% top-1 accuracy, NLL 0.8045, and ECE 0.0542. Thus the selected ensemble remains within the preservation budget while improving both NLL and ECE under the shared temperature. The same $r=2$ perturbations without correction obtain 80.619\% accuracy and NLL 0.9048.

The role of correction becomes more pronounced as perturbation magnitude increases. At $r=1$, corrected and uncorrected ID accuracy are nearly indistinguishable; at $r=2$ the corrected model begins to preserve accuracy better, and at $r=4$ the corrected ensemble loses 0.960 percentage points while the uncorrected ensemble loses 10.754 points. This scale dependence is consistent with the intended role of the affine repair: small perturbations require little correction, whereas larger perturbations expose a regime in which the repair is necessary to retain the base model's ID behavior.

\subsection{OOD detection and matched ablation}
\label{app:imagenet-ood}

Table~\ref{tab:imagenet_vit} reports the frozen-checkpoint comparison. At the primary operating point, P\&C obtains Near/Far AUROC $76.49/87.89$, compared with $72.53/84.40$ for the identical perturbations without correction and $73.52/86.04$ for MSP. Thus correction improves the matched perturbation ensemble by 3.96 Near-AUROC points and 3.49 Far-AUROC points; without correction, the $r=2$ perturbation ensemble is worse than MSP on both aggregates.

LLLA-Kron obtains Near/Far AUROC $72.95/87.42$, while a restricted linear-scope SCOD comparison obtains $75.15/88.48$. P\&C is therefore stronger than LLLA-Kron on Near OOD and slightly stronger on Far, and is stronger than SCOD-linear on Near while weaker on Far. A class-conditional Mahalanobis detector on the final normalized CLS representation remains stronger for pure OOD ranking, with Near/Far AUROC $78.82/92.55$. This detector leaves the predictive distribution unchanged, whereas P\&C constructs an ensemble predictive distribution. We therefore interpret the comparison as evidence that direct representation-space unsupportedness can be a stronger OOD score on this benchmark, rather than as evidence that the two methods estimate the same uncertainty object.

Across the strict, primary, and relaxed preservation budgets, P\&C Near/Far AUROC is $75.59/87.38$, $76.49/87.89$, and $77.27/88.06$, respectively. OOD ranking therefore improves as more of the allowed ID-preservation budget is used over the tested range. We retain the 0.50-percentage-point budget as the primary result rather than selecting the relaxed configuration from OOD performance.

\subsection{Representation geometry}
\label{app:imagenet-geometry}

The stronger Mahalanobis result is not explained simply by the rank of the P\&C perturbation subspace. Holding the same preservation protocol fixed, Near-OOD AUROC is $76.53$, $76.49$, $76.51$, and $76.41$ for $K\in\{5,20,40,80\}$. A random 20-dimensional projection of the final CLS representation also gives substantially weaker Mahalanobis performance than the full representation.

Instead, the methods emphasize different parts of the final representation geometry. Mahalanobis separation is strongest in low-variance ID directions: using the lowest-variance 96 covariance modes gives Near-OOD AUROC $81.81$, compared with $68.32$ for the highest-variance 96 modes. P\&C's induced final-CLS response has the opposite spectral weighting. Across covariance eigenmodes, response energy has Spearman correlation $0.993$ with the corresponding ID eigenvalue; 26.7\% of response energy lies in the highest-variance 96 modes and 4.69\% in the lowest-variance 96. This provides a concrete explanation for why representation distance can flag inputs that remain comparatively stable across the P\&C members.

\subsection{Construction and inference cost}
\label{app:imagenet-efficiency}

For the primary $M=20$ construction, the complete post-hoc procedure takes 284.7 seconds from a loaded base checkpoint and prepared data manifests. Of this, 276.3 seconds is the one-time feature-cache pass; constructing the member-specific perturbations and corrections after caching takes 8.4 seconds. Peak GPU memory is 1.49~GiB. The base checkpoint plus compact P\&C representation occupies approximately 510~MiB, compared with a derived 6.61~GiB for 20 complete ViT checkpoints.

The final-block target also allows almost the entire transformer prefix to be shared at inference. On the same GPU, the base ViT evaluates at 100.4 images/s and the cached-prefix $M=20$ P\&C ensemble at 98.4 images/s. The member-dependent tail still scales as $O(M)$; the small measured overhead arises because only the final feed-forward block and classifier-side computation must be evaluated separately for each member.

%% file: appendix_distilbert.tex
\section{Additional DistilBERT transfer experiment}
\label{app:distilbert}

We additionally apply P\&C inside the final feed-forward block of a fine-tuned 67M-parameter DistilBERT Banking77 classifier. This setting introduces self-attention, residual connections, LayerNorm, and categorical text prediction without retraining the checkpoint.

\input{tables/distilbert_tables}

Internal-FFN P\&C preserves $92.5\%$ ID accuracy with NLL $0.301$ and ECE $0.031$. Applying the same perturbations without affine correction retains $92.1\%$ accuracy but degrades NLL to $1.079$ and ECE to $0.525$, showing that the repair remains important in this architecture. On CLINC-OOS, P\&C is competitive with MC Dropout, while Energy remains the strongest OOD detector in this setting. We use this experiment as an additional test of whether the same post-hoc construction can operate inside a different model family rather than as a comprehensive language OOD benchmark.

%% file: tables/distilbert_tables.tex
\begin{table}[htbp]
\centering
\small
\caption{Banking77 ID prediction and CLINC-OOS detection with a 67M-parameter DistilBERT checkpoint. All methods use the same fine-tuned checkpoint. Internal-FFN P\&C and Uncorrected perturbation are averaged over five post-hoc construction seeds $\{0,10,42,123,2026\}$; MC Dropout is averaged over the same five stochastic inference seeds. Energy is deterministic for the fixed checkpoint.}
\label{tab:distilbert}
\begin{tabular}{lrrrrrr}
\toprule
Method & ID Acc. & ID NLL & ID ECE & Near AUROC & Cross AUROC & Far AUROC \\
\midrule
Energy & 0.925 & 0.303 & 0.032 & \textbf{0.929} & \textbf{0.980} & \textbf{0.990} \\
MC Dropout & 0.925 & \textbf{0.283} & \textbf{0.013} & 0.917 & 0.969 & 0.982 \\
Uncorrected perturb. & 0.921 & 1.079 & 0.525 & 0.885 & 0.972 & 0.989 \\
Internal-FFN P\&C & 0.925 & 0.301 & 0.031 & 0.907 & 0.967 & 0.981 \\
\bottomrule
\end{tabular}
\end{table}

%% file: main.bib
@inproceedings{ritter2018scalable,
  title={A scalable laplace approximation for neural networks},
  author={Ritter, Hippolyt and Botev, Aleksandar and Barber, David},
  booktitle={International conference on learning representations},
  year={2018}
}

@article{maddox2019simple,
  title={A simple baseline for bayesian uncertainty in deep learning},
  author={Maddox, Wesley J and Izmailov, Pavel and Garipov, Timur and Vetrov, Dmitry P and Wilson, Andrew Gordon},
  journal={Advances in neural information processing systems},
  volume={32},
  year={2019}
}

@article{osband2023epistemic,
  title={Epistemic neural networks},
  author={Osband, Ian and Wen, Zheng and Asghari, Seyed Mohammad and Dwaracherla, Vikranth and Ibrahimi, Morteza and Lu, Xiuyuan and Van Roy, Benjamin},
  journal={Advances in Neural Information Processing Systems},
  volume={36},
  pages={2795--2823},
  year={2023}
}

@article{lakshminarayanan2017simple,
  title={Simple and scalable predictive uncertainty estimation using deep ensembles},
  author={Lakshminarayanan, Balaji and Pritzel, Alexander and Blundell, Charles},
  journal={Advances in neural information processing systems},
  volume={30},
  year={2017}
}

@inproceedings{izmailov2020subspace,
  title={Subspace inference for Bayesian deep learning},
  author={Izmailov, Pavel and Maddox, Wesley J and Kirichenko, Polina and Garipov, Timur and Vetrov, Dmitry and Wilson, Andrew Gordon},
  booktitle={Uncertainty in Artificial Intelligence},
  pages={1169--1179},
  year={2020},
  organization={PMLR}
}

@article{dauphin2014identifying,
  title={Identifying and attacking the saddle point problem in high-dimensional non-convex optimization},
  author={Dauphin, Yann N and Pascanu, Razvan and Gulcehre, Caglar and Cho, Kyunghyun and Ganguli, Surya and Bengio, Yoshua},
  journal={Advances in neural information processing systems},
  volume={27},
  year={2014}
}

@inproceedings{choromanska2015loss,
  title={The loss surfaces of multilayer networks},
  author={Choromanska, Anna and Henaff, Mikael and Mathieu, Michael and Arous, G{\'e}rard Ben and LeCun, Yann},
  booktitle={Artificial intelligence and statistics},
  pages={192--204},
  year={2015},
  organization={PMLR}
}

@article{belkin2019reconciling,
  title={Reconciling modern machine-learning practice and the classical bias--variance trade-off},
  author={Belkin, Mikhail and Hsu, Daniel and Ma, Siyuan and Mandal, Soumik},
  journal={Proceedings of the National Academy of Sciences},
  volume={116},
  number={32},
  pages={15849--15854},
  year={2019},
  publisher={National Academy of Sciences}
}

@article{fawzi2018adversarial,
  title={Adversarial vulnerability for any classifier},
  author={Fawzi, Alhussein and Fawzi, Hamza and Fawzi, Omar},
  journal={Advances in neural information processing systems},
  volume={31},
  year={2018}
}

@article{d2022underspecification,
  title={Underspecification presents challenges for credibility in modern machine learning},
  author={D'Amour, Alexander and Heller, Katherine and Moldovan, Dan and Adlam, Ben and Alipanahi, Babak and Beutel, Alex and Chen, Christina and Deaton, Jonathan and Eisenstein, Jacob and Hoffman, Matthew D and others},
  journal={Journal of Machine Learning Research},
  volume={23},
  number={226},
  pages={1--61},
  year={2022}
}

@inproceedings{goodfellow2014explaining,
  title     = {Explaining and Harnessing Adversarial Examples},
  author    = {Goodfellow, Ian J. and Shlens, Jonathon and Szegedy, Christian},
  booktitle = {International Conference on Learning Representations},
  year      = {2015}
}

@inproceedings{gal2016dropout,
  title={Dropout as a bayesian approximation: Representing model uncertainty in deep learning},
  author={Gal, Yarin and Ghahramani, Zoubin},
  booktitle={international conference on machine learning},
  pages={1050--1059},
  year={2016},
  organization={PMLR}
}

@article{liu2020simple,
  title={Simple and principled uncertainty estimation with deterministic deep learning via distance awareness},
  author={Liu, Jeremiah and Lin, Zi and Padhy, Shreyas and Tran, Dustin and Bedrax Weiss, Tania and Lakshminarayanan, Balaji},
  journal={Advances in neural information processing systems},
  volume={33},
  pages={7498--7512},
  year={2020}
}

@inproceedings{madry2017towards,
  title={Towards Deep Learning Models Resistant to Adversarial Attacks},
  author={Madry, Aleksander and Makelov, Aleksandar and Schmidt, Ludwig and Tsipras, Dimitris and Vladu, Adrian},
  booktitle = {International Conference on Learning Representations},
  year      = {2018}
}

@article{hendrycks2016baseline,
  title={A baseline for detecting misclassified and out-of-distribution examples in neural networks},
  author={Hendrycks, Dan and Gimpel, Kevin},
  journal={arXiv preprint arXiv:1610.02136},
  year={2016}
}

@article{liu2020energy,
  title={Energy-based out-of-distribution detection},
  author={Liu, Weitang and Wang, Xiaoyun and Owens, John and Li, Yixuan},
  journal={Advances in neural information processing systems},
  volume={33},
  pages={21464--21475},
  year={2020}
}

@article{lee2018simple,
  title={A simple unified framework for detecting out-of-distribution samples and adversarial attacks},
  author={Lee, Kimin and Lee, Kibok and Lee, Honglak and Shin, Jinwoo},
  journal={Advances in neural information processing systems},
  volume={31},
  year={2018}
}

@article{sun2021react,
  title={React: Out-of-distribution detection with rectified activations},
  author={Sun, Yiyou and Guo, Chuan and Li, Yixuan},
  journal={Advances in neural information processing systems},
  volume={34},
  pages={144--157},
  year={2021}
}

@techreport{krizhevsky2009learning,
  title       = {Learning Multiple Layers of Features from Tiny Images},
  author      = {Krizhevsky, Alex and Hinton, Geoffrey},
  institution = {Computer Science Department, University of Toronto},
  address     = {Toronto, Ontario, Canada},
  year        = {2009},
  type        = {Technical Report}
}

@article{zhang2023openood,
  title={{OpenOOD} v1.5: Enhanced benchmark for out-of-distribution detection},
  author={Zhang, Jingyang and Yang, Jingkang and Wang, Pengyun and Wang, Haoqi and Lin, Yueqian and Zhang, Haoran and Sun, Yiyou and Du, Xuefeng and Li, Yixuan and Liu, Ziwei and others},
  journal={arXiv preprint arXiv:2306.09301},
  year={2023}
}

@inproceedings{todorov2012MuJoCo,
  title={MuJoCo: A physics engine for model-based control},
  author={Todorov, Emanuel and Erez, Tom and Tassa, Yuval},
  booktitle={2012 IEEE/RSJ international conference on intelligent robots and systems},
  pages={5026--5033},
  year={2012},
  organization={IEEE}
}

@article{towers2024gymnasium,
  title={Gymnasium: A standard interface for reinforcement learning environments},
  author={Towers, Mark and Kwiatkowski, Ariel and Terry, Jordan and Balis, John U and De Cola, Gianluca and Deleu, Tristan and Goul{\~a}o, Manuel and Kallinteris, Andreas and Krimmel, Markus and KG, Arjun and others},
  journal={arXiv preprint arXiv:2407.17032},
  year={2024}
}

@inproceedings{he2016identity,
  title={Identity mappings in deep residual networks},
  author={He, Kaiming and Zhang, Xiangyu and Ren, Shaoqing and Sun, Jian},
  booktitle={European conference on computer vision},
  pages={630--645},
  year={2016},
  organization={Springer}
}

@inproceedings{sharma2021sketching,
  title={Sketching curvature for efficient out-of-distribution detection for deep neural networks},
  author={Sharma, Apoorva and Azizan, Navid and Pavone, Marco},
  booktitle={Uncertainty in artificial intelligence},
  pages={1958--1967},
  year={2021},
  organization={PMLR}
}

@inproceedings{daxberger2021bayesian,
  title={Bayesian deep learning via subnetwork inference},
  author={Daxberger, Erik and Nalisnick, Eric and Allingham, James U and Antor{\'a}n, Javier and Hern{\'a}ndez-Lobato, Jos{\'e} Miguel},
  booktitle={International Conference on Machine Learning},
  pages={2510--2521},
  year={2021},
  organization={PMLR}
}

@article{wen2020batchensemble,
  title={Batchensemble: an alternative approach to efficient ensemble and lifelong learning},
  author={Wen, Yeming and Tran, Dustin and Ba, Jimmy},
  journal={arXiv preprint arXiv:2002.06715},
  year={2020}
}

@inproceedings{durasov2021masksembles,
  title={Masksembles for uncertainty estimation},
  author={Durasov, Nikita and Bagautdinov, Timur and Baque, Pierre and Fua, Pascal},
  booktitle={2021 IEEE/CVF Conference on Computer Vision and Pattern Recognition (CVPR)},
  pages={13534--13543},
  year={2021},
  organization={IEEE}
}

@article{dosovitskiy2020image,
  title={An image is worth 16x16 words: Transformers for image recognition at scale},
  author={Dosovitskiy, Alexey and Beyer, Lucas and Kolesnikov, Alexander and Weissenborn, Dirk and Zhai, Xiaohua and Unterthiner, Thomas and Dehghani, Mostafa and Minderer, Matthias and Heigold, Georg and Gelly, Sylvain and others},
  journal={arXiv preprint arXiv:2010.11929},
  year={2020}
}

@inproceedings{deng2009imagenet,
  title={Imagenet: A large-scale hierarchical image database},
  author={Deng, Jia and Dong, Wei and Socher, Richard and Li, Li-Jia and Li, Kai and Fei-Fei, Li},
  booktitle={2009 IEEE conference on computer vision and pattern recognition},
  pages={248--255},
  year={2009},
  organization={Ieee}
}
